\pdfoutput=1

\documentclass[11pt]{article}

\usepackage[final]{acl}
\usepackage{multicol}
\usepackage{times}
\usepackage{latexsym}

\usepackage[T1]{fontenc}

\usepackage[utf8]{inputenc}

\usepackage{microtype}

\usepackage{inconsolata}

\usepackage{graphicx}

\usepackage{times}
\usepackage{latexsym}

\usepackage{enumitem}
\usepackage{url}
\usepackage{booktabs}
\usepackage{graphicx}
\usepackage{multirow}
\usepackage{amsmath} 
\usepackage{amssymb}
\usepackage{flexisym}
\usepackage{listings}
\usepackage{wrapfig} 
\usepackage{xcolor}
\usepackage{color}
\usepackage{mdframed}
\usepackage[T1]{fontenc}
\usepackage{tcolorbox}
\usepackage{longtable}
\usepackage{colortbl}
\usepackage[utf8]{inputenc}

\usepackage{microtype}

\usepackage{inconsolata}

\usepackage{graphicx}

\usepackage{amsthm}

\theoremstyle{definition}
\newtheorem{proper}{Property}
\theoremstyle{remark}
\usepackage{adjustbox}
\usepackage{pifont}
\newcommand{\cmark}{\ding{51}}%
\newcommand{\xmark}{\ding{55}}%
\usepackage{threeparttable}
\newcommand{\benyou}[1]{{ \bf \color{red}   [benyou says `#1'] }}

\newcommand{\xuhan}[1]{{\bf \color{blue}  [xuhan says `#1']}}

\usepackage{amsthm}
\theoremstyle{plain}
\newtheorem{takeaway}{Take-away}

\usepackage{tikz}
\usetikzlibrary{mindmap,trees}
\definecolor{navy1}{rgb}{0.4, 0.5, 0.7}

\title{
LLMs for Mathematical Modeling: Towards Bridging the Gap between Natural and Mathematical Languages }

\definecolor{darkgreen}{rgb}{0.0, 0.5, 0.0}
\definecolor{maroon}{rgb}{0.5, 0.0, 0.0}
\definecolor{navy}{rgb}{0.0, 0.0, 0.5}

\definecolor{teal}{rgb}{0.0, 0.5, 0.5}
\author{
Xuhan Huang$^{1}$, Qingning Shen$^{1}$, Yan Hu$^{1}$\thanks{Corresponding author: \texttt{huyan@cuhk.edu.cn}}, Anningzhe Gao$^{2}$, Benyou Wang$^{1}$  \\
1 The Chinese University of Hong Kong, Shenzhen \\
2 Shenzhen Research Institute of Big Data \\
\texttt{xuhanhuang,qingningshen@link.cuhk.edu.cn, huyan@cuhk.edu.cn} \\
\texttt{anningzhegao@gmail.com, wangbenyou@cuhk.edu.cn} \\
}

\begin{document}

\maketitle

\begin{abstract}
Large Language Models (LLMs) have demonstrated strong performance across various natural language processing tasks, yet their proficiency in mathematical reasoning remains a key challenge. Addressing the gap between natural and mathematical language requires advanced reasoning capabilities, approaching those of Artificial General Intelligence (AGI). However, the evaluation remains challenging, as perfectly representing reality is inherently elusive, and traditional methods like manual or direct comparison of mathematical statements~\citep{ramamonjison_nl4opt_2023} are insufficient for assessing true modeling ability. We propose a process-oriented framework to evaluate LLMs' ability to construct mathematical models, using solvers to compare outputs with ground truth. Introducing \textbf{Mamo}, a benchmark with 1,209 questions covering ordinary differential equations, linear programming, and mixed-integer linear programming, we enable automatic evaluation of modeling accuracy. The results show that existing LLMs struggle with complex mathematical modeling tasks, with larger models demonstrating superior performance, while open-source models remain competitive in simpler cases but still fall short of proprietary models in more challenging problems.

\end{abstract}

\section{Introduction}




Recent advancements in Large Language Models (LLMs) have garnered widespread interest, demonstrating remarkable capabilities across a broad spectrum of natural language processing tasks \citep{ouyang2022training,openai2023gpt, nijkamp2022codegen,tang2024mathscale}. However, the domain of mathematical reasoning remains a critical aspect of the LLM evaluation \citep{wei2022chain}. This focus gauges LLMs' proficiency in solving mathematical challenges and reveals their underlying abilities in abstract reasoning and logic.

\textit{Natural language} reflects the complexity and subtlety of human communication, while \textit{mathematical language} provides accuracy and is ideal for representing the natural world \citep{firstCourseModeling}. Bridging their gap requires high-level cognitive skills akin to Artificial General Intelligence (AGI), a task at which LLMs show promise.
Recent studies have already combined LLMs with solvers \citep{feng2023language, pan2023logic}. In the field of optimization, \citet{ahmaditeshnizi2023optimus} introduced OptiMUS, an LLM-based agent that uses a solver to address optimization problems. The key task for LLM in OptiMUS involves first generating a mathematical formulation of a real-world problem, followed by writing Python code to invoke a solver. This implies the potential for testing LLM’s modeling capabilities. However, the challenge lies in the evaluation of modeling, as perfectly representing reality may be inherently elusive. 

To address this, we propose a new benchmarking strategy that evaluates mathematical modeling by utilizing solvers and answer verification. The core philosophy is to use solvers to resolve mathematical models and assess their accuracy by comparing the solver's output with the ground truth. Additionally, we introduce \textbf{Mamo}, a novel benchmark designed to assess the mathematical modeling capabilities of LLMs, leveraging solvers to shift the focus from conventional outcome-based evaluations to a more process-oriented approach. This strategy is particularly rational because it allows us to isolate and scrutinize the entire modeling process, from formulation to solution, providing a more holistic and rigorous evaluation of an LLM’s mathematical reasoning abilities.

The contributions of this paper are as follows:

 \begin{itemize}
     \item We introduce an automatic evaluation framework to assess mathematical modeling through \textit{solvers}, enabling exact answer matching. 

     \item We have developed a new benchmark, named \textbf{Mamo}, specifically tailored for the evaluation of mathematical modeling capabilities. Mamo encompasses a wide array of modeling questions (with a total of 1,209 meticulously curated questions), including ordinary differential equations (ODEs) and optimization problems within linear programming and mixed-integer linear programming frameworks. 
    
 \end{itemize}

\section{An Evaluation Framework for Mathematical Modeling}
\subsection{Definition: Mathematical Modeling}


Mathematical modeling is widely considered an advanced skill typically possessed by experts. According to \citet{firstCourseModeling}, mathematical models act as intermediaries that convert real-world problems into structured mathematical forms, facilitating the discovery of solutions. As illustrated in Figure \ref{fig:diff_M}, we aim to map a natural language question $Q$ into a mathematical model $M$, which is expressed in mathematical language as follows:
\begin{equation*}
\small
    f: Q \rightarrow M.
\end{equation*}

By solving this model, we can obtain an answer $A$ to the original problem $Q$.
Traditional mathematical modeling is a labor-intensive, highly specialized process that depends heavily on human expertise.

\subsection{Gap between Natural and Mathematical Languages}








\textit{Natural language} captures the complexity and nuance of human thought, while \textit{mathematical language} excels in precision and in modeling natural phenomena \citep{firstCourseModeling}. A single statement in natural language may correspond to different mathematical representations, and conversely, distinct natural language statements may map to the same mathematical formulation.

For example, consider the natural language statement: \textit{The car starts with an initial velocity of zero.} This can correspond to at least two different mathematical representations:

\textbf{Mathematical formulation 1}: $y'(0) = 0$

\textbf{Mathematical formulation 2}: $v(0)=0$.




Conversely, these two mathematical expressions may also correspond to natural language 2: \textit{The population growth rate is initially zero.}

Bridging the gap between \textit{natural language} and \textit{mathematical language} requires advanced cognitive abilities, a task where LLMs have shown significant potential. However, evaluating the effectiveness of mathematical modeling remains challenging, as achieving a perfect representation of reality through models is often unattainable.

                             

\begin{figure}[ht]
\centering
\includegraphics[width=0.35\textwidth]{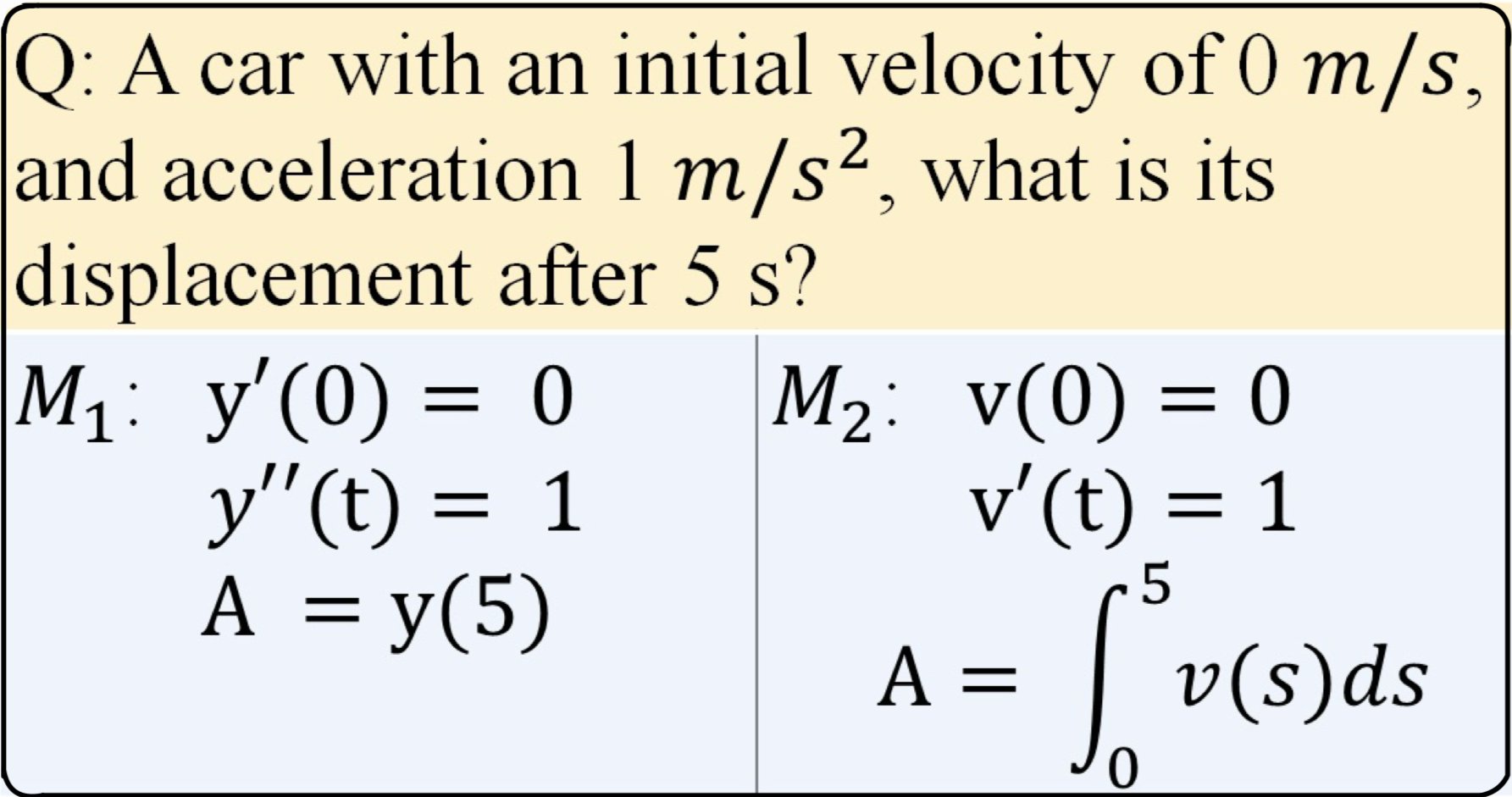}
\caption{$Q$ is the natural language problem, while $M_1$,$M_2$ are different mathematical models.}
\label{fig:diff_M}
\end{figure}

\paragraph{Challenges in Evaluation} Due to the complex mapping between \textit{natural language} and \textit{mathematical language}, a real-world problem $Q$ may correspond to multiple equivalent mathematical models, such as $M_1, M_2, M_3, \dots$. As illustrated in Figure \ref{fig:diff_M}, for a given problem $Q$, we have two distinct mathematical models, $M_1$ and $M_2$, both of which correctly describe the same real-world problem and are considered equivalent in the context of $Q$. Traditionally, the evaluation of mathematical models, including determining their equivalence, has been performed by human experts, a process that is both labor-intensive and time-consuming. In optimization modeling area, \citet{ramamonjison_nl4opt_2023} directly compare mathematical statements by standardizing variable names and evaluating accuracy based on declaration matches. While this approach addresses differences in variable naming, the abstraction of variables from natural language context, which is critical in mathematical modeling, was not examined.

\begin{figure*}[ht]
\centering
\includegraphics[width=1.0\textwidth]{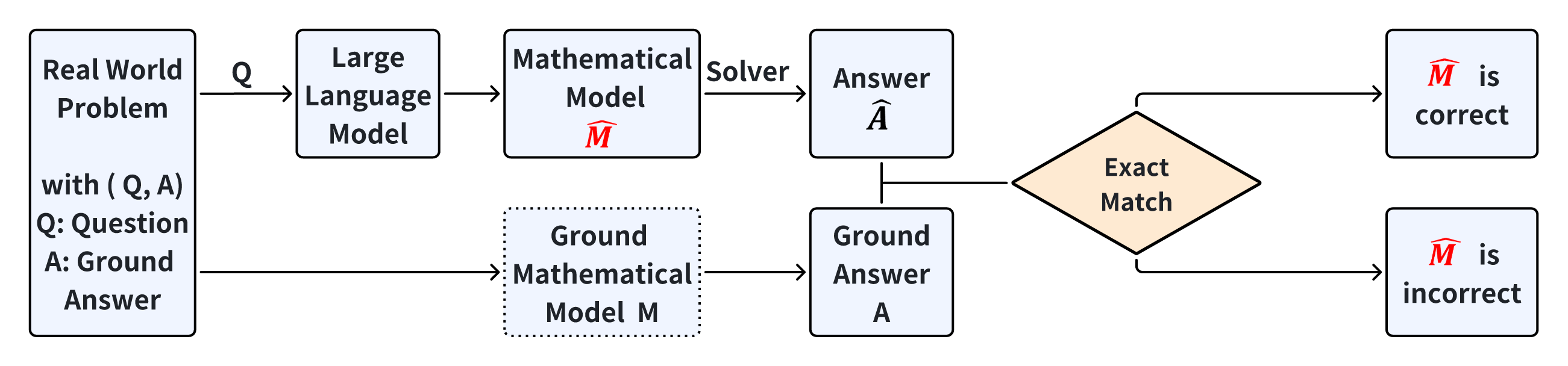}
\caption{The pipeline to use exact answer verification via an additional solver.}
\label{fig:methodology}
\end{figure*}

\subsection{Evaluating Modeling Using Solvers and  Answer Matching } 
\label{sec:solvers}

\paragraph{The Evaluation Framework}
To facilitate the \textit{automatic} evaluation of mathematical models, we propose an evaluation framework for mathematical modeling based on exact answer verification using solvers, as illustrated in Figure \ref{fig:methodology}. A solver is an algorithmic tools designed to find solutions to various classes of problems. For instance, Gurobi is a well-known optimization solver \citep{gurobi}, while Python packages \citep{virtanen2020scipy} and MATLAB \citep{MATLAB} are commonly used for solving ODEs. Solvers are essential for operationalizing abstract models, denoted as ${\color{red}\widehat{M}}$, generated by large language models (LLMs), translating them into concrete solutions $\widehat{A}$ that can be analyzed and validated against real-world data.
For a given real-world problem and its corresponding answer pair $(Q, A)$, LLMs generate a mathematical model, denoted $\textcolor{red}{\widehat{M}}$. The solver resolves this model, producing a solution $\widehat{A}$, which is then compared to the correct answer $A$. This framework directly evaluates the LLM's problem-solving capability, with any inaccuracies in $\color{red}\widehat{M}$ typically leading to incorrect solutions $\widehat{A}$. By focusing on the final output, we avoid additional computational steps that could obscure the evaluation of the LLM's performance.

\paragraph{The Philosophy}
The philosophy behind our evaluation framework is \textbf{\textit{goal-driven}}, based on the idea that the ultimate aim of mathematical modeling is to produce accurate solutions to real-world problems. We prioritize the correctness of the solution $\widehat{A}$ over the intermediate steps of model formulation. By comparing the solver's output with the correct answer, we evaluate the effectiveness of the LLM's generated model $\color{red}\widehat{ M}$. The assumption here is that if the model is well-constructed, it will lead to an accurate solution. 

Despite its \textit{\textbf{goal-driven}} structure, the framework retains a \textbf{\textit{process-oriented}} emphasis on the modeling capability of LLMs. Unlike traditional benchmarks that treat answer correctness as an isolated metric, our approach explicitly evaluates the intermediate step of model 
 formulation. The evaluative focus lies on whether the LLM constructs a mathematical model $\color{red} \widehat{M}$ that faithfully translates the real-world problem into mathematical language. Thus, while the solver’s answer validates the outcome, the core objective remains to assess the \textit{modeling process}. 
 
 This dual perspective allows us to maintain a scalable yet principled evaluation framework. While the solver’s output remains the validation mechanism, the primary objective is to assess the LLM’s ability to conceptualize and formulate mathematical problems rather than simply solving predefined mathematical tasks (more discussions are provided in Appendix~\ref{sec:correct_ans_wrong_model}).

\section{Mamo Benchmark}
To implement this framework, we developed a new benchmark, \textbf{Mamo}, specifically for mathematical modeling.
\label{sec:philosophy}
\subsection{Scope of Mamo Benchmark}
\label{sec:scope_of_modeling}

Given the broad spectrum of mathematical modeling~\cite{firstCourseModeling} (see Appendix~\ref{sec:scope}) and the availability of specific solvers, our benchmark is meticulously designed to encompass domains of ODEs and linear programming problems. This section outlines the rationale behind our focus on these areas, driven by the criteria of \textit{solver availability} and \textit{result verifiability}.

\paragraph{Availability of Sophisticated Solvers:} Advanced optimization solvers, such as COPT \citep{copt}, along with Python libraries for solving ODEs, offer a strong foundation for testing the capabilities of LLMs. These tools enable a detailed and automated evaluation of the LLMs' abilities to abstract, formulate, and accurately solve mathematical problems within these specific domains.


\paragraph{The Straightforward Verifiability of  Results}~
While calculators address basic problems in these areas, the absence of a structured modeling approach hinders the assessment of LLMs' ability to create solvable mathematical models. 

Building on these considerations, we developed a benchmark focused on optimization and ODE, given the lack of universal PDE solvers. Compared to NLP4LP, introduced by \cite{ahmaditeshnizi2023optimus} and focusing exclusively on optimization, our benchmark includes a larger number of problems, with potential expansions into nonlinear optimization, probabilistic modeling, and PDE as new solvers emerge.

\subsection{Data Selection}


Our benchmark dataset comprises manually selected and GPT-generated questions, curated by a team with advanced mathematical education, including Ph.D. holders, and an average GPA of 3.89/4.0, ensuring strong analytical expertise. Details on the collectors' qualifications are provided in Appendix \ref{sec:annotatorQualification}.

\paragraph{Source  Reliability} 

For manual selection, we referenced textbooks and solutions in ODE and optimization~\citep{Thomas_Weir_Hass_Giordano_2016, Stewart_Clegg_Watson_2014, Lial_Greenwell_Ritchey_2017, Braun_1993, Boyce_DiPrima_2012, Giordano_Horton_Fox_2014, Zill_2013, Bertsimas_Tsitsiklis_1997, Hurlbert_2010}. We focused on deriving mathematical models and assigning realistic scenarios, ensuring answer accuracy. Each question was adapted to meet the benchmark criteria (Section \ref{sec:criteria}) by adjusting scenarios or solutions as needed.

\paragraph{Question Formulation and Verification}
Data synthesis starts by creating mathematical constructs with random parameters, followed by answer computation to establish ground truth. {\tt GPT-4} then generates real-life scenarios, translating these models into context-rich problems, testing both mathematical reasoning and real-world contextualization. See the example in Appendix \ref{sec:data_synthesis}.


\subsection{Guidance of Data Collection}\label{sec:criteria}

\begin{figure}[ht]
\centering
\includegraphics[width=0.5\textwidth]{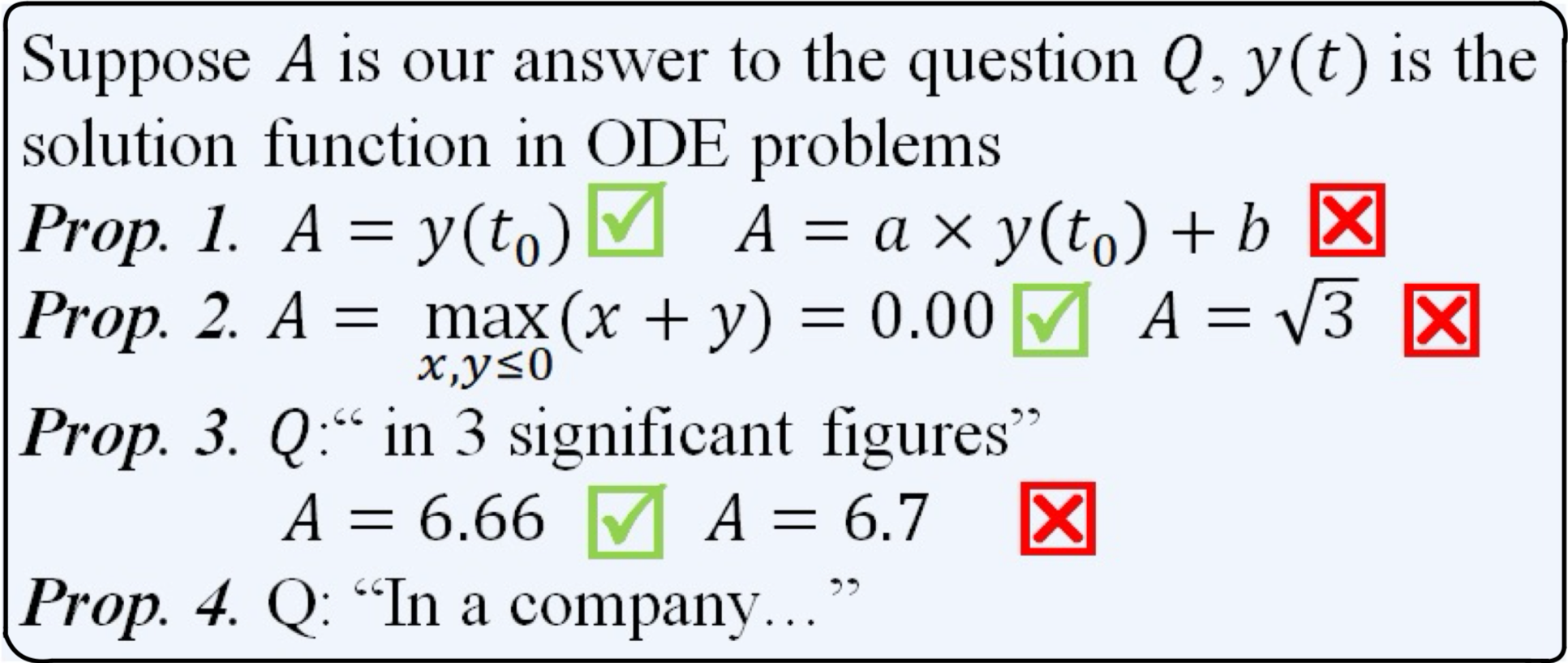}
\caption{$Q$ is the natural language problem (may be ODE problem or optimization problem), where y is a solution function in ODE problems.}
\label{fig:example_prop}
\end{figure}

\begin{table*}[h!]
\small
\centering
\begin{tabular}{lcccc}
\toprule
\textbf{Category} & \textbf{Initial Questions} & \textbf{Filtered}&  \textbf{Corrected}& \textbf{Final Questions} \\ 
\midrule
ODE          & 383                         & 37        &41        & 346                       \\ 
Easy LP      & 688                         & 36        &~~8        & 652                       \\ 
Complex LP   & 211                         & ~~0         &40       & 211                       \\ \bottomrule
\end{tabular}
\caption{Summary of the review process.}
\label{tab:review_summary}
\end{table*}

Our benchmark dataset follows specific criteria to ensure the problems are suitable for automated verification. Figure~\ref{fig:example_prop} provides examples of these criteria, with further explanations presented in Appendix~\ref{sec:explanation}.

\begin{proper}{\bf Final-State Approach Solvability~} 
\label{proper:1}
Problems should be structured so that a solver can directly compute the solution once the LLM formulates the mathematical model. For example, asking for the optimal value of an optimization problem is solvable using a 'final-state approach', whereas transforming an ODE solution requires extra steps and is not solvable this way.

\end{proper}    

\begin{proper}{\bf Unified and Numerical Answers~} 
    
For ODEs, questions should seek the function's value at a specific time \(t_0\), and for optimization problems, the optimal value, ensuring a unified and numerical answer for clear comparison with the solver's output.
\label{proper:2}
\end{proper}

\begin{proper}{\bf Significant Figures and Precision~}

Each question will explicitly state the required significant figures or the level of numerical precision for the answer. 
\label{proper:3}
\end{proper}

To assess the LLM's capability to abstract mathematical models from real-world scenarios, we introduce Property~\ref{proper:4}.

\begin{proper}{\bf Real-World Problem Context}
Questions should be framed as real-world problems without explicitly stating the underlying mathematical model.
 \label{proper:4}
\end{proper}

\begin{table*}[ht!]
\small
\centering
\begin{minipage}{0.48\textwidth}
    \centering
    \begin{tabular}{lcccc}
        \toprule
        \textbf{Reviewer} & Reviewer 1 & Reviewer 2 & Reviewer 3 & Reviewer 4\\ 
        \midrule
        Reviewer 1 & -   & 0.67  & 0.71  & 0.56  \\ 
        Reviewer 2 & 0.67 & -    & 0.65  & 0.50  \\ 
        Reviewer 3 & 0.71 & 0.65 & -    & 0.52  \\ 
        Reviewer 4 & 0.56 & 0.50 & 0.52  & -    \\
        \bottomrule
    \end{tabular}
    \caption{Inter-reviewer Cohen’s Kappa scores.}  
    \label{tab:review_summary_kappa}
\end{minipage}
\hfill
\begin{minipage}{0.48\textwidth}
    \centering
    \footnotesize
    \begin{tabular}{lc}
        \toprule
        \textbf{Reviewer} & \textbf{Accuracy Rate (\%)} \\ 
        \midrule
        Reviewer 1  & 90.0 \\ 
        Reviewer 2  & 84.0 \\ 
        Reviewer 3  & 88.0 \\ 
        Reviewer 4  & 74.0 \\ 
        \bottomrule
    \end{tabular}
    \caption{Accuracy of reviewers.}
    \label{tab:reviewer_accuracy}
\end{minipage}
\end{table*}
\subsection{Data Quality Checking}

In the ODE section, we utilized {\tt GPT-4} to verify the solvability of generated questions as stated in Property~\ref{proper:1}, filtering out any invalid questions. A detailed evaluation identified issues such as missing information, incorrect answers, and unclear phrasing, leading to the removal or correction of problematic questions and resulting in a refined dataset. In the optimization section, specifically in the Easy\_LP part, invalid questions were eliminated, and those with incorrect answers or misleading descriptions were corrected. For the Complex\_LP section, questions involving incorrect mathematical models were revised (see an example in Appendix~\ref{sec:example_correction}). A summary of the review process is provided in Table~\ref{tab:review_summary}.

For the cross-review process, 50 questions from the ODE dataset were selected for in-depth analysis by four independent reviewers~\footnote{The hourly wage is approximately 20 US dollars, with total compensation of around 473 US dollars.}. Reviewers were chosen based on their qualifications, including an average GPA of at least 3.5/4.0 in fundamental math courses, completion or enrollment in the Ordinary Differential Equation course, and meeting English proficiency standards.



Reviewers evaluated the questions, classifying responses as either numerical answers or "error" if the question was problematic. The effectiveness of this review was assessed using Cohen's Kappa to measure inter-rater reliability, along with individual accuracy rates. As shown in Table~\ref{tab:review_summary_kappa}, the pairwise Cohen's Kappa scores had an average of 0.60, reflecting moderate to substantial agreement among reviewers, with individual scores ranging from 0.50 to 0.71. Accuracy rates were moderately high, as detailed in Tables~\ref{tab:reviewer_accuracy}. For additional details on the review, see Appendix \ref{sec:crossReview}.

\subsection{Data Statistics}
\label{sec:datadescription}
In the section on Ordinary Differential Equations, we present a total of 346 problems:~196 are based on first-order equations, 110 on second-order equations, and 40 on systems of equations, offering a comprehensive exploration across different levels. The optimization section is divided into two segments: Easy\_LP, featuring 652 high school-level Mixed Integer Linear Programming (MILP) problems, and Complex\_LP, comprising 211 undergraduate-level problems that integrate both LP and MILP.
See the category statistics and the word cloud of the combined data in Appendix~\ref{sec:wordCloud}.

\section{Evaluation}

\subsection{Evaluation Protocol}

\subsubsection{Evaluation Details}
\label{sec:eval_detail}
Our evaluation protocol follows the methodology outlined in Section~\ref{sec:solvers}. First, the LLMs were prompted to generate formalized code (in each domain) to represent a mathematical model, denoted as $\color{red} \widehat{M}$. This code was then executed by a solver, and the result, $\widehat{A}$, was compared with the ground truth answer, $A$.

For ODE problems, the LLM was prompted to generate Python code based on a natural language description. The generated code was executed, and the output was compared to the correct solution to assess its accuracy. For optimization problems, the LLM was asked to express the model in {\tt .lp} format, after which an optimization solver COPT~\citep{copt} 
was employed to find the optimal value for comparison with the correct answer. This process is independent of the specific solver used; for instance, MATLAB~\citep{MATLAB} can be used as an ODE solver, and Gurobi~\citep{gurobi} can serve as an alternative optimization solver.

\paragraph{Benchmarking Models}
Our evaluation\footnote{See the settings in Appendix~\ref{sec:setting}} includes both proprietary and open-source LLMs. The proprietary LLMs consist of the GPT-4 series~\citep{openai2023gpt}, the Claude series\footnote{\href{https://www.anthropic.com/news/claude-3-family}{https://www.anthropic.com/news/claude-3-family}}, and the Gemini series\footnote{\href{https://deepmind.google/technologies/gemini/}{https://deepmind.google/technologies/gemini/}}. The open-source LLMs include the DeepSeek series~\citep{deepseek-math, deepseekv2}, Llama-3.1 models~\citep{dubey2024llama3herdmodels}, Qwen-2.5~\citep{qwen2.5}, and Mixtral~\citep{jiang2024mixtral}.


\paragraph{Model Output} We prompted the LLMs to generate Python code that called solvers (such as {\tt solve\_ivp} and {\tt odeint} in SciPy, and {\tt dsolve} in NumPy). In the optimization sections, the LLMs were asked to generate a standard {\tt .lp} file. Compared to the testing methodology in NLP4LP~\citep{ahmaditeshnizi2023optimus}, which involved prompting the model to output Python code for invoking the optimization solver, the {\tt .lp} format more closely aligned with the standard optimization form, making it more readable. Optimization solvers like COPT~\citep{copt} could directly read and solve the {\tt .lp} file. To better adhere to this format, we conducted tests using 3-shot learning. Detailed information about the number of few-shot samples used in our experiments and the results can be found in Appendix~\ref{sec:few_shot_experiment} and Appendix~\ref{sec:prompt}. Additionally, at the start of the testing process, we performed an initial cleaning of the code, removing strings such as \texttt{ \`{}\`{}\`{} python \`{}\`{}\`{}} that would have affected our testing.




\subsubsection{Evaluation Metrics}
\label{subsec:metric}
According to Property~\ref{proper:2} and Property~\ref{proper:3}, both the ground truth answer $A$ and the answer $\widehat{A}$, derived from solving the LLM's output mathematical model $\color{red}\widehat{M}$, are numerical values with specified significant figures. To account for minor discrepancies, we apply the following comparison rule:
\paragraph{General Comparison Criterion}: Let \(n\) denote the number of decimal places in \(A\) (\(n = 0\) if \(A\) is an integer). The answer \(\widehat{A}\) is considered correct if:
$$\left| \frac{\widehat{A} - A}{A} \right| < 10^{-4} \text{ or } |\widehat{A} - A| < \min\{10^{-n},10^{-2}\}.$$

\begin{figure}[ht]
\centering
\includegraphics[width=0.5\textwidth]{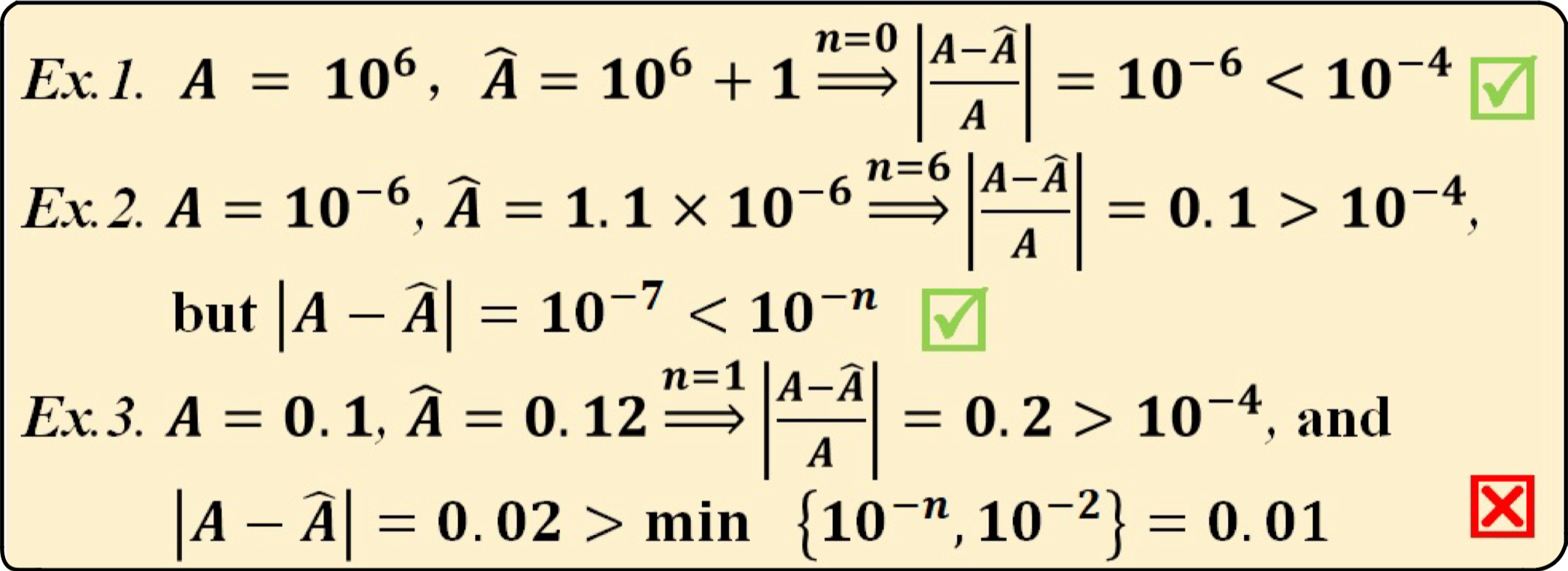}
\caption{Examples illustrating the application of the General Comparison Criterion}
\label{fig:example_comp}
\end{figure}


This rule accounts for precision errors and ensures an accurate evaluation of $\widehat{A}$ while allowing for slight computational variations. Figure~\ref{fig:example_comp} demonstrates the application of this criterion with examples, illustrating how discrepancies between the correct and generated answers are assessed. For further details, see Appendix~\ref{sec:evaluation_script} and the case study in Appendix~\ref{sec:metric}.




\subsection{Benchmarking Results}
\label{sec:benchmark_result}
\begin{table*}
\centering
\footnotesize
\begin{threeparttable}
    \begin{tabular}{l|lll|ll|r}
    \toprule
    \multirow{3}{*}{\textbf{Models}} & \multicolumn{3}{c|}{\textbf{ODE}} & \multicolumn{2}{c|}{\textbf{Optimization}} & \multirow{3}{*}{\textbf{Overall~(\%)}}\\
    \cline{2-6}
    &\scalebox{0.87}{\textbf{\raggedright First}} & \scalebox{0.87}{\textbf{Second}}& \scalebox{0.87}{\textbf{System}}&\scalebox{0.87}{\textbf{Easy}} &\scalebox{0.87}{\textbf{Complex}} &  \\
    &\scalebox{0.87}{\textbf{order~(\%)}} & \scalebox{0.87}{\textbf{order~(\%)}}& \scalebox{0.87}{\textbf{(\%)}}&\scalebox{0.87}{\textbf{LP~(\%)}} &\scalebox{0.87}{\textbf{LP~(\%)}} \\
    \midrule
   \multicolumn{7}{c}{ Proprietary Models}\\
    o1-preview & \underline{\bf71.43 \scalebox{0.7}{\color{darkgreen} +0.00}} & \underline{{\bf50.00 \scalebox{0.7}{\color{darkgreen} +0.00}}} & \textbf{45.00 \scalebox{0.7}{\color{darkgreen} +0.00}} & 80.21 \scalebox{0.7}{\color{darkgreen} +0.00} & \underline{{\bf36.02 \scalebox{0.7}{\color{darkgreen} +3.79}}} & 67.60 \scalebox{0.7}{\color{darkgreen} +0.66} \\
    GPT-4o & {67.86 \scalebox{0.7}{\color{darkgreen} +0.51}} & 36.36 \scalebox{0.7}{\color{darkgreen} +0.91} & 40.00 \scalebox{0.7}{\color{darkgreen} +0.00} & 87.27 \scalebox{0.7}{\color{darkgreen} +0.15} & 22.75 \scalebox{0.7}{\color{darkgreen} +8.53} & 66.67 \scalebox{0.7}{\color{darkgreen} +1.73} \\
    GPT-4-turbo & 64.80 \scalebox{0.7}{\color{darkgreen} +1.02} & 32.73 \scalebox{0.7}{\color{darkgreen} +0.91} & 40.00 \scalebox{0.7}{\color{darkgreen} +0.00} & 87.88 \scalebox{0.7}{\color{darkgreen} +0.00} & 23.22 \scalebox{0.7}{\color{darkgreen} +6.64} & 66.34 \scalebox{0.7}{\color{darkgreen} +1.41} \\
    GPT-4 & 59.18 \scalebox{0.7}{\color{darkgreen} +1.02} & 28.18 \scalebox{0.7}{\color{darkgreen} +0.91} & 40.00 \scalebox{0.7}{\color{darkgreen} +2.50} & 86.50 \scalebox{0.7}{\color{darkgreen} +0.46} & 21.08 \scalebox{0.7}{\color{darkgreen} +3.09} & 63.81 \scalebox{0.7}{\color{darkgreen} +1.12} \\
    GPT-3.5-turbo & 16.33 \scalebox{0.7}{\color{darkgreen} +6.12} & ~~7.27 \scalebox{0.7}{\color{darkgreen} +4.55} & 10.00 \scalebox{0.7}{\color{darkgreen} +5.00} & 81.29 \scalebox{0.7}{\color{darkgreen} +3.53} & ~\:9.48 \scalebox{0.7}{\color{darkgreen} +0.00} & 49.13 \scalebox{0.7}{\color{darkgreen} +3.48} \\
    Claude-3-opus & 55.61 \scalebox{0.7}{\color{darkgreen} +3.57} & 33.64 \scalebox{0.7}{\color{darkgreen} +1.81} & {\bf 45.00 \scalebox{0.7}{\color{darkgreen} +2.50}} & 83.74 \scalebox{0.7}{\color{darkgreen} +0.31} & ~\:9.48 \scalebox{0.7}{\color{darkgreen} +0.47} & 60.38 \scalebox{0.7}{\color{darkgreen} +1.08} \\
    Claude-3-sonnet & 52.04 \scalebox{0.7}{\color{darkgreen} +2.04} & 24.55 \scalebox{0.7}{\color{darkgreen} +0.90} & 27.50 \scalebox{0.7}{\color{darkgreen} +5.00} & 83.59 \scalebox{0.7}{\color{darkgreen} +0.61} & 24.64 \scalebox{0.7}{\color{darkgreen} +0.48} & 60.96 \scalebox{0.7}{\color{darkgreen} +0.99} \\
    Claude-3-haiku & 41.33 \scalebox{0.7}{\color{darkgreen} +9.18} & 23.64 \scalebox{0.7}{\color{darkgreen} +0.91} & 17.50 \scalebox{0.7}{\color{darkgreen} +2.50} & 78.53 \scalebox{0.7}{\color{darkgreen} +7.36} & 17.54 \scalebox{0.7}{\color{darkgreen} +1.42} & 54.84 \scalebox{0.7}{\color{darkgreen} +5.87} \\
    Gemini-1-pro & 25.00 \scalebox{0.7}{\color{darkgreen} +8.16} & 13.64 \scalebox{0.7}{\color{darkgreen} +0.00} & \;\;2.50 \scalebox{0.7}{\color{darkgreen} +2.50} & 78.83 \scalebox{0.7}{\color{darkgreen} +1.08} & 14.69 \scalebox{0.7}{\color{darkgreen} +0.00} & 50.45 \scalebox{0.7}{\color{darkgreen} +1.99} \\
    Gemini-1.5-pro & 64.80 \scalebox{0.7}{\color{darkgreen} +1.02} & 30.00 \scalebox{0.7}{\color{darkgreen} +0.00} & {\underline{\bf 45.00 \scalebox{0.7}{\color{darkgreen} +5.00}}} & 85.28 \scalebox{0.7}{\color{darkgreen} +0.00} & 30.81 \scalebox{0.7}{\color{darkgreen} +5.21} & 66.09 \scalebox{0.7}{\color{darkgreen} +1.24} \\
    \hline
    \multicolumn{7}{c}{ Open-source Models}\\
     DeepSeek-v2.5 &{67.86 \scalebox{0.7}{\color{darkgreen} +0.00}} & 36.36 \scalebox{0.7}{\color{darkgreen} +0.00} & 32.50 \scalebox{0.7}{\color{darkgreen} +0.00} & 86.20 \scalebox{0.7}{\color{darkgreen} +1.99} & 13.33 \scalebox{0.7}{\color{darkgreen} +0.00} & 65.27 \scalebox{0.7}{\color{darkgreen} +1.07} \\
    DeepSeek-math-7B-base &  ~\:3.57 \scalebox{0.7}{\color{darkgreen} +0.00} & ~\:2.73 \scalebox{0.7}{\color{darkgreen} +0.00} & ~\:2.50 \scalebox{0.7}{\color{darkgreen} +0.00} & 49.39 \scalebox{0.7}{\color{darkgreen} +1.22} & ~\:6.64 \scalebox{0.7}{\color{darkgreen} +0.00} & 28.70 \scalebox{0.7}{\color{darkgreen} +0.66} \\
    DeepSeek-math-7B-rl & ~\:1.02 \scalebox{0.7}{\color{darkgreen} +0.51} & ~\:1.82 \scalebox{0.7}{\color{darkgreen} +0.00} & ~\:5.00 \scalebox{0.7}{\color{darkgreen} +0.00} & 20.55 \scalebox{0.7}{\color{darkgreen} +0.62} & ~\:0.00 \scalebox{0.7}{\color{darkgreen} +0.47} & 11.58 \scalebox{0.7}{\color{darkgreen} +0.50} \\

    Llama-3.1-8B-instruct & 30.61 \scalebox{0.7}{\color{darkgreen} +1.53} & ~\:4.55 \scalebox{0.7}{\color{darkgreen} +1.81} & 10.00 \scalebox{0.7}{\color{darkgreen} +2.50} & 74.39 \scalebox{0.7}{\color{darkgreen} +2.60} & 13.27 \scalebox{0.7}{\color{darkgreen} +0.00} & 48.14 \scalebox{0.7}{\color{darkgreen} +1.90} \\
     Llama-3.1-70B-instruct & 62.24 \scalebox{0.7}{\color{darkgreen} +0.52} & 33.64 \scalebox{0.7}{\color{darkgreen} +5.45} & 32.50 \scalebox{0.7}{\color{darkgreen} +0.00} & 85.74 \scalebox{0.7}{\color{darkgreen} +0.36} & 22.75 \scalebox{0.7}{\color{darkgreen} +1.42} & 64.44    \scalebox{0.7}{\color{darkgreen} +1.02} \\
      Llama-3.1-405B-instruct & 66.84 \scalebox{0.7}{\color{darkgreen} +1.02} &  40.00 \scalebox{0.7}{\color{darkgreen} +0.00} & 30.00 \scalebox{0.7}{\color{darkgreen} +0.00} & 86.20 \scalebox{0.7}{\color{darkgreen} +0.76} & 34.12 \scalebox{0.7}{\color{darkgreen} +4.74} &{\underline{\bf 67.91 \scalebox{0.7}{\color{darkgreen} +1.40}}}\\
    Qwen-2.5-32B-instruct & 60.71 \scalebox{0.7}{\color{darkgreen} +0.51} & 38.18 \scalebox{0.7}{\color{darkgreen} +0.00} & 35.00 \scalebox{0.7}{\color{darkgreen} +0.00} & 84.05 \scalebox{0.7}{\color{darkgreen} +0.00} & 12.32 \scalebox{0.7}{\color{darkgreen} +1.42} & 61.95 \scalebox{0.7}{\color{darkgreen} +0.33} \\
    Qwen-2.5-72B-instruct & 60.71 \scalebox{0.7}{\color{darkgreen} +0.00} & 37.27 \scalebox{0.7}{\color{darkgreen} +0.91} & 32.50 \scalebox{0.7}{\color{darkgreen} +0.00} & {\underline{\bf 89.42 \scalebox{0.7}{\color{darkgreen} +0.15}}} & 11.43 \scalebox{0.7}{\color{darkgreen} +1.90} & 64.53 \scalebox{0.7}{\color{darkgreen} +0.50} \\

    Mixtral-8x7B-instruct & 23.47 \scalebox{0.7}{\color{darkgreen} +4.59} & ~\:4.55 \scalebox{0.7}{\color{darkgreen} +1.81} & 10.00 \scalebox{0.7}{\color{darkgreen} +0.00} & 67.79 \scalebox{0.7}{\color{darkgreen} +2.15} & ~\:9.95 \scalebox{0.7}{\color{darkgreen} +0.00} & 42.85 \scalebox{0.7}{\color{darkgreen} +2.06} \\
    Mixtral-8x22B-instruct & 52.04 \scalebox{0.7}{\color{darkgreen} +1.53} & 11.82 \scalebox{0.7}{\color{darkgreen} +1.82} & 20.00 \scalebox{0.7}{\color{darkgreen} +0.00} & 81.90 \scalebox{0.7}{\color{darkgreen} +2.61} & 19.91 \scalebox{0.7}{\color{darkgreen} +1.42} & 57.82 \scalebox{0.7}{\color{darkgreen} +2.06} \\
    \bottomrule
    \end{tabular}
    \begin{tablenotes}
             \footnotesize
          \item [1] Here {\tt GPT-4-turbo} denotes {\tt GPT-4-turbo-2024-04-09}; {\tt GPT-4o} denotes {\tt GPT-4o-2024-05-13}; {\tt GPT-4} refers to {\tt GPT-4-0613}; {\tt GPT-3.5-turbo} denotes {\tt GPT-3.5-turbo-0125.} 
    \end{tablenotes}
\end{threeparttable}
\caption{Evaluation Results on the Mamo Benchmark. The {\color{darkgreen} subscript} denotes improvement using the original LLM as a \textbf{code modifier}, which corrects syntax errors without changing the underlying logic, differentiating modeling from formatting errors (see Section~\ref{para:modifier}). The "Overall" score represents the weighted average of correct rates, weighted by question count. The highest score in each category is marked in \textbf{boldface}; an \underline{{\color{darkgreen} underline}} signifies the top score with the LLM itself as code modifier. The score after using LLM as code modifier is computed as the \textbf{sum} of the original score and the value in the subscript. }
\label{tab:result_1}
\end{table*}

\begin{table*}
\centering
\vspace{-10pt}
\label{ExecutableRate}

\footnotesize
\setlength\tabcolsep{2.5pt}

\begin{tabular}{l|rrr|rrr}
\toprule
\multirow{2}{*}{\textbf{Models}} & \multicolumn{3}{c|}{\textbf{ODE} (\%)} & \multicolumn{3}{c}{\textbf{Optimization} (\%)}\\
 & {raw} & {self-modified} & {GPT-4-modified} & {raw} & {self-modified} & {GPT-4-modified} \\
\midrule
\multicolumn{7}{c}{ Proprietary Models}\\
o1-preview & 99.71 & 100.00 & 100.00 & 97.80 & 99.65 & 98.73\\
GPT-4o & 96.82 & 99.42 & 99.42 & 92.93 & 97.91 & 97.91\\
GPT-4-turbo & 96.82 & 99.13 & 98.84 & 93.86 & 97.80 & 97.56\\
GPT-4 & 94.51 & 97.69 & 97.69 & 90.96 & 95.13 & 95.13\\
GPT-3.5-turbo & 37.28 & 65.03 & 93.93 & 87.95 & 89.80 & 96.76\\
Claude-3-Sonnet & 84.68 & 93.06 & 97.40 & 91.89 & 93.97 & 96.76\\
Claude-3-Haiku & 78.03 & 92.77 & 98.27 & 82.61 & 92.58 & 96.29\\
Claude-3-Opus & 88.73 & 96.82 & 97.98 & 90.38 & 93.40 & 95.94\\
Gemini-1.0-pro & 56.36 & 69.94 & 88.73 & 88.06 & 89.11 & 94.21\\
Gemini-1.5-pro & 93.64 & 98.27 & 98.84 & 95.48 & 98.73 & 96.87\\
\hline
    \multicolumn{7}{c}{ Open-source Models}\\
DeepSeek-v2.5 & 97.69 & 99.13 & 99.13 & 90.36 & 93.83 & 97.55 \\
DeepSeek-math-7B-base & 28.90 & 29.77 & 79.77 & 55.50 & 58.40 & 75.67\\
DeepSeek-math-7B-rl & 5.20 & 6.64 & 80.64 & 19.93 & 21.55 & 55.97\\
Llama-3.1-8b-instruct & 57.51 & 72.97 & 92.20 & 80.99 & 88.30& 94.55 \\
Llama-3.1-70b-instruct & 92.48 & 98.27 & 97.98 & 92.93 & 96.99 & 97.34\\
Llama-3.1-405b-instruct & 98.84 & 99.71 & 99.71 & 92.58 & 98.03 &95.48\\
Qwen-2.5-32B-instruct & 97.69 & 98.84 & 99.13 & 93.17 & 96.29 & 94.67\\ 
Qwen-2.5-72B-instruct & 97.40 & 97.69 & 97.98 & 90.80 & 94.06 & 91.15 \\
Mixtral-8x7B-instruct & 50.87 & 69.08 & 91.62 & 84.13 & 87.83 & 96.29\\
Mixtral-8x22B-instruct & 80.92 & 88.73 & 96.24 & 87.72 & 93.51 & 96.29\\
\bottomrule

\end{tabular}
\caption{Execution Success Rates for Raw, Self-Modified, and GPT-4 Modified Methods, denoting scenarios without code modifiers, utilizing the LLM as a code modifier, and employing {\tt GPT-4-0613} as the code modifier. This table highlights the success rates of \textbf{file execution} (correct format) across these three scenarios, \text{differing} from the Table~\ref{tab:result_1} which shows the \textbf{correctness rate}.}
\label{table:result_improve}
\end{table*}

Table \ref{tab:result_1} presented the evaluation results (accuracy rate) of different language models on the Mamo Benchmark. The performance of these models was assessed both in their original form and when enhanced by the language model (i.e., code modifier) itself to rectify syntax errors (with improvements indicated by the {\color{darkgreen} subscript}). Appendix~\ref{sec:gpt_modifier} provided results when {\tt GPT-4-0613} was used as the code modifier, and Appendix~\ref{sec:error_analysis} presented the error analysis.

\paragraph{Mamo is Challenging} As the results in Table~\ref{tab:result_1} indicate, even state-of-the-art models struggled with complex tasks such as second-order ODEs, systems of ODEs, and Complex\_LP. This is due to the increasing number of variables and the more intricate relationships among these variables compared to easier tasks like Easy\_LP.

\begin{takeaway}
Existing LLMs struggle with complex tasks in mathematical modeling.
\end{takeaway}

\paragraph{The Performance of o1} Interestingly, {\tt o1-preview}, although it excelled in complex tasks such as second-order ODEs and Complex\_LP, did not outperform (and was even worse than {\tt GPT-3.5-Turbo}) in Easy\_LP tasks, which consist of high-school-level linear programming modeling problems. As shown in Table~\ref{tab:result_2}, the execution success rates of {\tt o1-preview} in each category were nearly 100\% across all methods (raw, self-modified, or GPT-4-modified), indicating that this underperformance was not due to limitations in coding ability.

\begin{takeaway}
{\tt o1} excelled in complex tasks but underperformed in easy tasks.
\end{takeaway}

\begin{figure*}[ht!]
\centering
\includegraphics[width=1.0\textwidth]{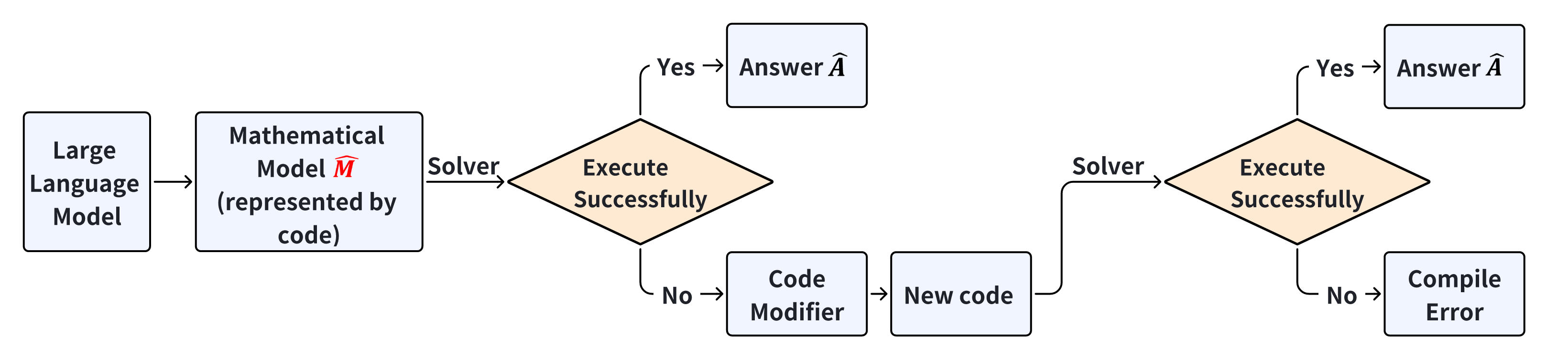}
\caption{The testing pipeline involving code modifier. }
\label{fig:pipeline_code_modifier}
\end{figure*}

\paragraph{Open-source Models vs. Proprietary Models} One of the most notable aspects of this benchmark is the emergence of open-source models as serious competitors to closed-source models. The {\tt Llama-3.1-405B}, in particular, achieved the best overall score across both evaluation categories. The {\tt Llama-3.1} series, {\tt Qwen} series, and {\tt DeepSeek-v2.5}, despite being open-source models, demonstrated remarkable performance, even surpassing highly regarded commercially developed models such as GPT-4 and Claude-3 in categories like second-order ODEs and Optimization.

\begin{takeaway}
Open-source models are competitive in simpler tasks, but there is still a gap compared to closed-source models in more complex tasks.
\end{takeaway}

\paragraph{Effectiveness of Scaling} The evaluation demonstrates a correlation between model size and performance across almost all tested categories. As indicated by the results of the {\tt Llama-3.1}, {\tt Qwen}, and {\tt Mixtral} series, for models with the same architecture but different parameter scales (e.g., {\tt Llama-3.1-7B} vs. {\tt Llama-3.1-405B}), larger models tended to outperform their smaller counterparts. This observation highlights the effectiveness of scaling in the mathematical reasoning tasks of LLMs. However, the effectiveness of scaling may plateau on easier tasks, such as Easy\_LP problems.

\begin{takeaway}
Larger LLMs perform better in mathematical modeling.
\end{takeaway}

\subsection{On the Impact of Format Errors}

\paragraph{Mitigation of Formatting Errors}
\label{para:modifier}
It is important to differentiate between errors in \textit{modeling} and those related to coding or file \textit{formatting} (See Appendix~\ref{sec:error_format}). When the LLM output contains syntax errors in Python or {\tt .lp} code, these errors must be corrected without altering the logic of the code to ensure that the evaluation focuses solely on modeling. To address syntax issues, we employ a \textbf{code modifier}, as the pipeline shown in Figure~\ref{fig:pipeline_code_modifier} (More details are discussed in Appendix~\ref{sec:under_code_modifier}). In our experiments, we used the original language model and {\tt GPT-4-0613} as code modifiers, with the former selected for \textit{reproducibility} and the latter for its superior coding \textit{proficiency} and \textit{stability}. A comparison of results from different code modifiers is provided in Section~\ref{sec:benchmark_result} and Appendix~\ref{sec:gpt_modifier}, with prompts detailed in Appendix~\ref{sec:prompt}.

\paragraph{Setting}~
The testing pipeline is shown in Figure~\ref{fig:pipeline_code_modifier}. As mentioned in Section~\ref{sec:eval_detail}, we selected the original LLM and {\tt GPT-4-0613} as code modifiers. The data presented in Table \ref{table:result_improve} provides a comparison of the performance (rate of correct format) of various models using different modification methods: i) \textit{no modification}, ii) \textit{self-modification}, and iii) \textit{modification provided by {\tt GPT-4-0613}}.
\paragraph{Observation}
Combining Table~\ref{tab:result_1} and Table~\ref{table:result_improve}, the {\color{darkgreen} improvement} can be attributed to misjudgments caused by the LLM's coding ability. For LLMs with a lower initial correct format rate (such as {\tt GPT-3.5-Turbo} and {\tt Llama-3.1-8B} in ODE tasks), the relative improvement is higher. This is because, in several cases, their output codes, initially judged as "Error" due to compile errors, actually represent correct mathematical models. It is also worth noting that the {\color{darkgreen} improvement} has little effect on the ranking, indicating that the influence of different coding abilities on our benchmark is minimal.

\begin{takeaway}
Format errors have minimal impact on mathematical modeling.
\end{takeaway}

\section{Related Work}

Recent studies have advanced the use of large language models (LLMs) in mathematical problem-solving by introducing new datasets and solver methodologies. For example, \citet{frieder2024mathematical} and \citet{yuan2023well} developed datasets to test LLMs on various mathematical problems and arithmetic expressions. \citet{lewkowycz2022solving} focused on training models using natural language paired with LaTeX-formatted math from arXiv. To address complex computational tasks, \citet{zhang2024evaluating} introduced the CARP dataset, while \citet{he2024olympiadbench} presented OlympiadBench, a bilingual benchmark for competition-level math reasoning. \citet{liu2024mathbench} also developed MathBench, structured to assess LLMs' theoretical knowledge. Additionally, \citet{mirzadeh2024gsmsymbolicunderstandinglimitationsmathematical} proposed GSM-Symbolic, a benchmark that generates symbolic variants of GSM8K questions. Their study also introduced GSM-NoOp, a dataset containing irrelevant but seemingly related information to assess model robustness, demonstrating that while LLMs exhibit sensitivity to numerical variations, they remain resilient to superficial modifications.

In solver development, efforts have focused on integrating LLMs into novel problem-solving approaches. \citet{yang2023large} explored LLMs as direct solvers, highlighting their autonomous problem-solving potential. \citet{he2023solving} introduced a hybrid method that combines LLMs with external symbolic solvers for equation solving, while \citet{pan2023logic} developed LOGIC-LM, a framework blending LLMs with symbolic solvers to tackle logical problems. \citet{ahmaditeshnizi2023optimus} contributed to both dataset development and solver integration by creating NLP4LP, a benchmark for LP and MILP problems, and OptiMUS, an LLM-based agent for optimization problem solving. In line with these developments, \citet{lyu2023faithfulchainofthoughtreasoning} introduced Faithful-CoT, a two-stage framework that translates natural language into symbolic reasoning chains (represented as code) before solving the problem using corresponding solvers. This approach aligns closely with our benchmark’s objective of assessing the faithful translation of problems into structured mathematical formulations.

\section{Conclusion and Future Directions}

Mathematical modeling is a crucial yet challenging aspect of mathematical reasoning. Evaluating LLMs' ability to handle modeling is essential, as their performance will continue to improve over time, making specialized benchmarks in specific areas necessary to accurately assess and push their capabilities. Due to the complexity of mathematical models, assessment is difficult. To address this, we introduce the Mamo benchmark, which incorporates solvers to validate answers and assess the correctness of the modeling process. Mamo demonstrates that LLMs can be effectively evaluated on complex reasoning tasks, where the correct answer may vary but remains equivalent in principle. This approach paves the way for further research into LLMs' reasoning capabilities.



\clearpage
\section*{Limitations}
\label{sec:limit}
While the current benchmarking methodology necessitates LLMs to generate Python code for ODEs and {\tt .lp} files for optimization problems, it opens up exciting avenues for future research. The process, while effective, may inadvertently influence the LLMs' focus towards formalization over the conceptual modeling, potentially affecting the depth of mathematical reasoning. This opens the door for benchmarks that better distinguish between an LLM's modeling skills and formalization abilities. Future work could focus on assessing LLMs' mathematical modeling directly, bypassing formalization for a more nuanced evaluation.

\section*{Ethics Statement:} 
\label{sec:ethics}
The benchmark is designed with a stringent ethical framework to ensure that all problems are socially responsible and do not perpetuate or encourage harmful biases or stereotypes. Questions are meticulously reviewed to avoid any content that could lead to the dissemination of misinformation, support unethical practices, or cause harm to individuals or groups. This includes but is not limited to issues of privacy, security, and the potential for misuse of the model. Furthermore, the benchmark abstains from any problem that could indirectly endorse unethical behaviors or decisions in real-world scenarios, upholding the highest standards of academic integrity and social responsibility.

\section*{Acknowledgment}
This work was supported by  the Shenzhen Science and Technology Program (JCYJ20220818103001002), Shenzhen Doctoral Startup Funding (RCBS20221008093330065), Tianyuan Fund for Mathematics of National Natural Science Foundation of China (NSFC) (12326608), Shenzhen Key Laboratory of Cross-Modal Cognitive Computing (grant number ZDSYS20230626091302006), and Shenzhen Stability Science Program 2023.
\newpage

\bibliography{naacl_2025}

\clearpage
\appendix
\section{{\tt GPT-4-0613} as Code Modifier}
\label{sec:gpt_modifier}

\renewcommand{\arraystretch}{0.9}

\begin{table*}[h]
\centering

\begin{threeparttable}
\begin{tabular}{l|lll|ll|r}
\toprule
\multirow{3}{*}{\textbf{Models}} & \multicolumn{3}{c|}{\textbf{ODE}} & \multicolumn{2}{c|}{\textbf{Optimization}} & \multirow{3}{*}{\textbf{Overall~(\%)}}\\
\cline{2-6}
&\textbf{ \raggedright First} & \textbf{Second}& \textbf{System}&\textbf{Easy} &\textbf{Complex} &  \\
&\textbf{order~(\%)} & \textbf{order~(\%)}& \textbf{(\%)}&\textbf{LP~(\%)} &\textbf{LP~(\%)} \\
\midrule
\multicolumn{7}{c}{ Proprietary Models}\\
    o1-preview $^{\dagger\dagger}$ & {\textbf{71.43}} & \textbf{50.00} & 45.00  & 80.21 & \textbf{37.91}  & 67.49\\
    GPT-4o $^{\dagger\dagger}$ & {68.37} & 37.27 & 40.00  & 87.42 & 29.38  & 68.07\\
    GPT-4-turbo $^{\dagger\dagger}$ & 65.82 & 33.64 & 40.00  & 88.19 & 27.96  & 67.58 \\
    GPT-4 $^{\dagger\dagger}$ & 60.20 &	29.09 &	42.50  & 86.96 & 24.17 & 64.93\\
    GPT-3.5-turbo $^{\dagger\dagger}$ & 44.90 & 20.00 & 20.00 & 85.43 & ~\:9.48 & 57.48 \\
    Claude-3-Opus $^{\dagger\dagger}$ & 59.18 & 36.36 & 47.50 & 84.20 & 14.22 & 62.37\\
    Claude-3-Sonnet $^{\dagger\dagger}$ & 56.63 & 26.36 & 32.50& 85.12 & 27.96 & 63.44 \\
    Claude-3-Haiku $^{\dagger\dagger}$ & 51.53 & 25.45 & 22.50 & 86.81 & 18.96 & 61.54 \\
    Gemini-1-pro $^{\dagger\dagger}$ & 40.31 & 18.18 & 10.00 & 80.21 & 15.64 & 54.51 \\
    Gemini-1.5-pro $^{\dagger\dagger}$ & 65.82 & 30.91 & \textbf{50.00} &  85.28 & 34.12 & 67.08 \\
    \midrule
\multicolumn{7}{c}{ Open-source Models}\\
    DeepSeek-v2.5 $^{\dagger\dagger}$ & 67.86 & 36.36 & 32.50 & 86.20 & 13.33 & 64.20 \\
    DeepSeek-math-7b-base $^{\dagger\dagger}$ & 13.78 & ~~9.09 & ~~5.00  & 59.05 & ~~9.48  & 36.72 \\
    DeepSeek-math-7b-rl $^{\dagger\dagger}$ & 13.78 & ~~7.27 & ~~7.50 &  47.70 & ~~1.90 & 29.20 \\
    Llama-3.1-8b-instruct $^{\dagger\dagger}$  &33.67 & ~~8.18 & 15.00 & 82.06 & 13.27& 53.27\\
    Llama-3.1-70b-instruct $^{\dagger\dagger}$ & 62.76 & 39.09 & 32.50 & 86.81 & 23.70 & 65.76  \\
    Llama-3.1-405b-instruct $^{\dagger\dagger}$  & 67.86 & {40.00} & 30.00 & 86.66 & {34.60} & {\bf 68.40}\\
    Qwen-2.5-32b-instruct $^{\dagger\dagger}$ & 61.73 & 38.18 & 35.00 & 84.05 & 13.27 & 62.28 \\
     Qwen-2.5-72b-instruct $^{\dagger\dagger}$ & 60.71 & 37.27 & 32.50 & {\bf 89.57} & 11.43 & 64.61 \\
    Mixtral-8x7b-instruct $^{\dagger\dagger}$ & 32.14 & 11.82 & 20.00 & 75.46 & ~~9.95 & 49.38\\
    Mixtral-8x22b-instruct $^{\dagger\dagger}$ & 55.10 & 17.27 & 22.50 & 85.43 & 22.27 & 61.21 \\

\bottomrule
\end{tabular}
\end{threeparttable}
\caption{Evaluation result on Mamo Benchmark. The remark $\dagger\dagger$ indicates it additionally uses the {\tt GPT-4-0613} to rectify syntax error of the code and {\tt .lp} file generated by the corresponding LLM.}
\label{tab:result_2}
\end{table*}

Selecting GPT-4 as a code modifier generally results in better performance than using the LLM itself for self-improvement. For instance, DeepSeek-v2's accuracy in ODE tasks improves from 36.41\% with self-modification to 48.80\% when GPT-4 is used as the code modifier. This pattern indicates that GPT-4 is a more effective code modifier, enhancing the accuracy and performance of various models more reliably than self-improvement.

\section{Cross Review}


\label{sec:crossReview}
In ODE part, we have the following review process:
\begin{enumerate}
    \item We employed {\tt GPT-4} to confirm that the newly generated questions are amenable to being solved by solvers using the final-state approach. There were only 10 out of 383 questions are invalid. All were filtered, 373 questions were left.
    \item Then we conducted a thorough evaluation of the questions. Our dataset initially comprised \(373\) questions. We identified \(29\) questions that lacked sufficient information, \(21\) that had incorrect answers, and \(18\) that featured unclear statements. Following our assessment, \(27\) questions were deleted, and \(41\) were corrected to ensure clarity and correctness.
\end{enumerate}
Finally, we have 346 problems in ODE part.

We have also conducted the cross-review process. See the details in Appendix~\ref{sec:review_process} and the results on Appendix~\ref{sec:review_result}.

In the optimization section, particularly with easy LP part, out of 688 entries, 12 were found to be completely invalid (not meeting the criteria), 26 had incorrect answers, and six had misleading descriptions. All invalid entries were filtered out, and the remaining issues were corrected manually. Among them, eight of them were corrected and others (36 questions) were filtered out. In complex LP part, for the original 211 questions, 40 were found formulated by wrong mathematical models. All of them were corrected.


\subsection{Cross-Review Process}
\label{sec:review_process}
Furthermore, adhering to the scaling law evident in the dataset's distribution, we selected 50 questions from the ODE parts in our dataset for a deeper analysis. Four independent reviewers (see their qualification in Section~\ref{sec:annotatorQualification}) were tasked with assessing 50 questions each ~\footnote{The hourly wage is approximately 20 US dollars (equivalent in local currency), with a total participant compensation of around 473 US dollars (equivalent in local currency).}. The responses could either be numeric or labeled as ``error'' if a question was deemed problematic. This section outlines the metrics used to evaluate the reviewers' responses against pre-defined correct answers, which also include the ``error'' label for any flawed questions. We use the same metric as described in Subsection~\ref{subsec:metric}. Here are our results:

\subsubsection{Results}
\label{sec:review_result}
The effectiveness of the review process was quantified using the metrics defined in Subsection~\ref{subsec:metric}, focusing on the accuracy and inter-rater reliability among reviewers. The statistical outcomes are summarized as follows:

\begin{itemize}
    \item \textbf{Average Cohen's Kappa:} The average Cohen's Kappa across all reviewer pairs was 0.60, indicating a moderate to substantial agreement. This suggests that reviewers generally agreed on the classification of answers, though there were variations in some cases.
    \item \textbf{Minimum and Maximum Cohen's Kappa:} The minimum Kappa value recorded was 0.50, and the maximum was 0.71. The spread in these values highlights areas where alignment and training could potentially enhance consistency.
    \item \textbf{Accuracy Rates:} The individual accuracy rates were as follows:
    \begin{itemize}
        \item Reviewer 1: 90.0\%
        \item Reviewer 2: 84.0\%
        \item Reviewer 3: 88.0\%
        \item Reviewer 4: 74.0\%
    \end{itemize}
    These rates reflect the precision with which each reviewer matched the standard answers, including their recognition of problematic questions correctly labeled as ``error.''
\end{itemize}

These results provide insights into the overall effectiveness and areas of improvement for the cross-review process, particularly in terms of aligning understanding and interpretation of the evaluation criteria among reviewers.

\section{Collector Qualifications}
\label{sec:annotatorQualification}


The benchmarking process benefits from the expertise of reviewers with robust backgrounds in mathematics, underscored by their academic accomplishments. The team is composed of individuals whose education encompasses critical mathematical disciplines, including calculus (I and II), linear algebra (both introductory and advanced levels), optimization, probability, and ordinary differential equations. The average Grade Point Average (GPA) across these foundational courses is approximately 3.89 out of a 4.0 scale. This high level of academic achievement demonstrates the reviewers' strong grasp of essential mathematical concepts and their application. Furthermore, the team is enriched by members who hold Ph.D. degrees in mathematics, further solidifying the depth of expertise and analytical skills available for the benchmarking process.
\subsection{Qualification of Reviewers}
We select the reviewers mentioned in Appendix~\ref{sec:review_process} based on the following criteria:
\begin{enumerate}
\item Undergraduate students who have taken fundamental mathematical courses (Calculus and Linear Algebra) and have an average GPA of at least 3.5/4.0.
\item Have completed or are currently taking courses in Ordinary Differential Equations (ODE).
\item Meet one of the following English proficiency requirements: TOEFL > 80, IELTS > 6.5, a minimum grade of B+ in all English classes, or a Gaokao English score > 120.
\end{enumerate}
These criteria ensure their ability to read questions in English and solve ODE problems.

\section{Understand Code Modifier}
\label{sec:under_code_modifier}
Figure \ref{fig:pipeline_code_modifier} shows the evaluation process involving the code modifier. The introduction of code modifiers is to maintain the fairness of the evaluation: we try to reduce conditions when the LLM outputs the code with a correct mathematical model, but the code ends with a compile error due to the limit of coding ability. We concider two factors when applying code modifier:
\begin{enumerate}[label=(\alph*)]
    \item How often does a format error occur?
    \item How many passes on average are needed?
\end{enumerate}
The factor (a) is exactly the Table~\ref{table:result_improve}. As for (b), the number of passes required to fix format errors depends on the code modifier chosen. To study the factor (b), 
To study factor (b), we applied GPT-4-0613 as a code modifier on the chosen format error files and iteratively called it until the code was executed successfully. Table~\ref{tab:number_code_modifier} displays the distribution of the number of pass the code modifier was called until the errors were fixed. The chosen files were generated by GPT-4o and DeepSeek-v2.5, representing a close-source and an open-source model, respectively.

\begin{table}[ht]
\centering
\begin{tabular}{|c|c|c|c|c|c|c|}
\hline
     & \textbf{1} & \textbf{2} & \textbf{3} & \textbf{4} & \textbf{$\geq$5} & \textbf{all} \\
\hline
\textbf{LP}  & 74 & 35 & 10 & 6 & 11 & 136 \\
\textbf{ODE} & 88 & 3  & 1  & 1 & 9  & 112 \\
\hline
\end{tabular}
\caption{Number of pass to apply code modifier}
\label{tab:number_code_modifier}
\end{table}
On average, 1.68 passes were needed to correct format errors when using GPT-4-0613. We found that a single pass was sufficient to correct the majority of format errors. To balance efficiency and accuracy, we only applied a single pass for the code modifier throughout our experiments.

\section{Metric}
\label{sec:metric}
The comparison process is as the following:
\begin{equation*}
\text{$C_{A,\widehat{A}}$} = 
\begin{cases} 
1, & \begin{aligned}
       &\text{if } \left| \frac{\widehat{A} - A}{A} \right| \leq 1 \times 10^{-4} \text{~{\color{red} (1)}} \\
       &\text{or } \\
       & |\widehat{A} - A| < 1 \times \min\{10^{-n}, 10^{-2}\} \text{~{\color{red} (2)}},
    \end{aligned} \\ \\
0, & \text{otherwise.}
\end{cases}
\end{equation*}

Our consideration:
{\color{red} (1)} The difference is much less than the Answers, so can be omitted.\\
{\color{red} (2)} The only difference lies in the last few digits of the required significant figure of the answer. By selecting different thresholds \(\xi\times min\{10^{-n},10^{-2}\}\), we can adjust the tolerance for the discrepancy, ensuring that \(|\widehat{A} - A| < \xi\times min\{10^{-n},10^{-2}\}\). We have explored the relationship between various thresholds and the accuracy of GPT-4o.
\begin{table*}[h]
\centering
\caption{Accuracy at various thresholds for different categories(\%), where $\xi_i = i$}
\label{tab:diff_threshold}
\begin{tabular}{c|cccccccccc}
\toprule
$\xi$ & 0 & 1 & 2 & 3 & 4 & 5 & 6 & 7 & 8 & 9 \\
\midrule
ODE & 48.55 & 50.87 & 54.62 & 56.36 & 57.51 & 58.96 & 59.83 & 60.12 & 60.69 & 61.85 \\
Easy LP & 87.27 & 87.27 & 87.27 & 87.27 & 87.27 & 87.27 & 87.27 & 87.27 & 87.27 & 87.27 \\
Complex LP & 22.75 & 22.75 & 22.75 & 22.75 & 22.75 & 22.75 & 22.75 & 22.75 & 22.75 & 22.75 \\
\bottomrule
\end{tabular}
\end{table*}

\begin{figure*}[h]
\centering
\includegraphics[width=0.9\textwidth]{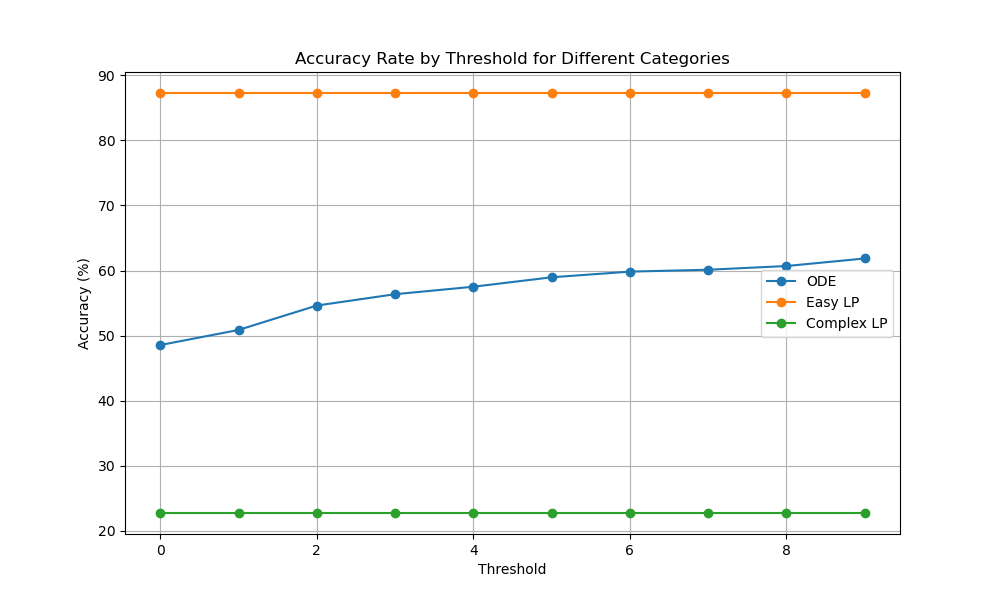}
\caption{Accuracy of different threshold in different category}
\label{fig:thershold}
\end{figure*}

Figure~\ref{fig:thershold} plots the results. In our metric, we choose strict $\xi = 1$.

\section{Error Analysis}


\label{sec:error_analysis}
In this section, we will analyze some typical errors.
\subsection{Deviation from Instruction}
It is common for some models that show a low capability of few-shot learning, especially for small models.  
For example, in a case Llama-3-8B output {\tt `I wish you good luck'} when prompting with some few-shot demonstrations in mathematical modeling. 



\subsection{Syntax Error}
\begin{figure*}[ht]
    \centering
    \begin{minipage}[b]{0.3\textwidth}
        \includegraphics[width=\textwidth]{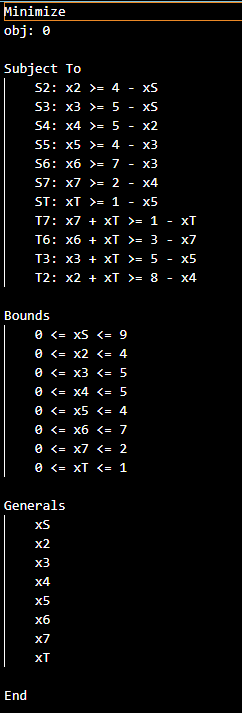}
        \caption{The syntax error is caused by the constraint, in .lp format, the variables should be placed on the left hand side of the inequality}
        \label{fig:lp_syntax_1}
    \end{minipage}
    \hfill 
    \begin{minipage}[b]{0.4\textwidth}
        \includegraphics[width=\textwidth]{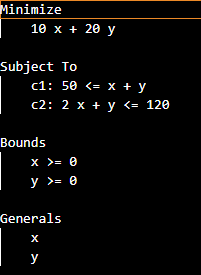}
        \caption{Same type of syntax error as Figure~\ref{fig:lp_syntax_1}}
        \label{fig:lp_syntax_2}
    \end{minipage}
\end{figure*}

In ODE parts, the syntax errors is obviously the syntax errors in the reponsed python code.
In Optimization part, the .lp file require more strict syntax than the universal optimization formulation. The Figure~\ref{fig:lp_syntax_1} and~\ref{fig:lp_syntax_2} show the typical syntax errors in .lp format.

\subsection{Modeling Error}
\begin{figure*}[ht]
\centering
\includegraphics[width=1\textwidth]{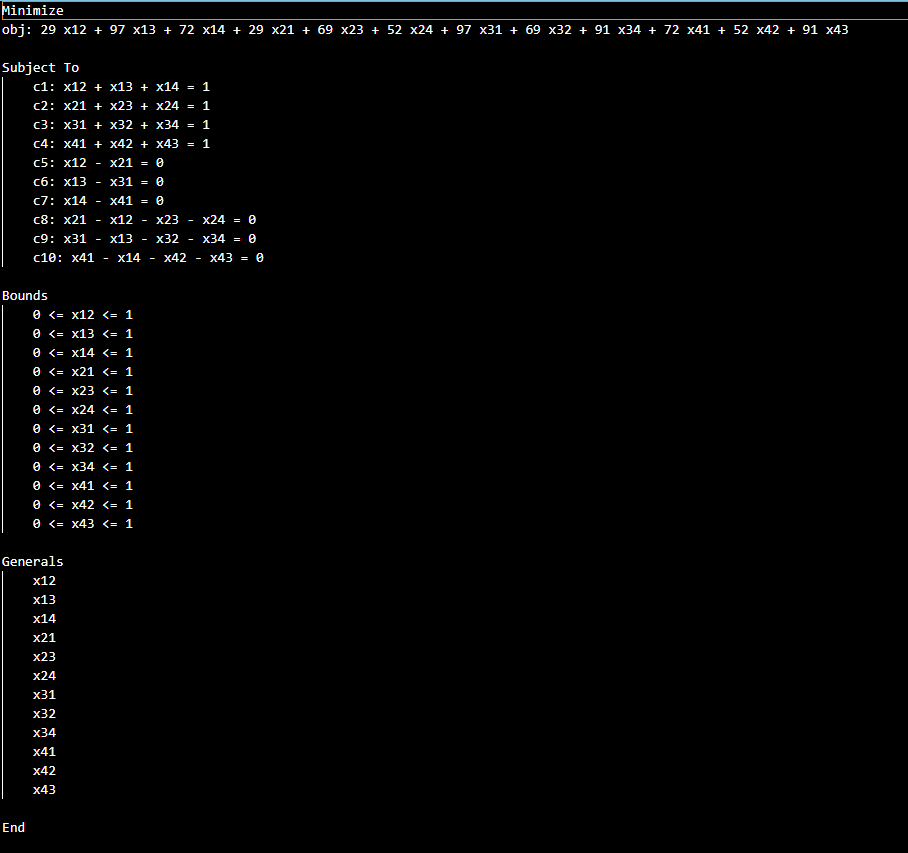}
\caption{In this TSP problem, the decision variable should be binary, instead of "General", which means ``integer''}
\label{fig:model_error}
\end{figure*}

One of the most common errors in modeling is the lack of recognition of decision variable types. The types of decision variables in optimization problems are divided into discrete (sometimes integer, sometimes binary) and continuous. Figure~\ref{fig:model_error}
is an example.

\section{Scope of Mamo}
\label{sec:scope}

\begin{table*}[ht]
\centering

\begin{tabular}{lc}
\toprule
Category &  In Mamo \\ \midrule
Changes and Differences  & \xmark \\ 
Proportionality and Geometric Similarity  & \xmark \\ 
Optimization Problem Modeling   & \cmark \\ 
Differential Equations   & \cmark \\ 
Probabilistic Modeling   & \xmark \\
\bottomrule

\end{tabular}
\caption{The taxonomy of mathematical modeling. In Mamo benchmark, we only include the categories where sophisticated Solvers is available.}
\label{table:taxonomy}
\end{table*}
The scope of Mamo is chosen under the consideration in Section~\ref{sec:scope_of_modeling}. As Table~\ref{table:taxonomy} shows, Mamo focus on the 
ODE and Optimization areas.
\section{Examples of Different ${\widehat{M}}$}
\label{sec:diff_model}
Different models will typically lead to different final answers after solving by the solvers. For example, in ODE part: \\
$${\widehat{M}}_{1}:y' + y = 0$$ and  $${\widehat{M}}_{2}: y' + 2y = 0$$
Suppose we ask for $y(1)$, then we have ${\widehat{A}}_1 = e^{-1}$ and ${\widehat{A}}_1 = e^{-2}$,they have different values.\\
Then for the optimization, \\
${\widehat{M}}_{1}$:
\begin{equation*}
\begin{aligned}
& \max_{x,y}
& & x + y \\
& \text{subject to}
& & x + y \leq 20, \\
& & & x - y \geq 0, \\
& & & 0 \leq x \leq 10, \\
& & & 0 \leq y \leq 5, \\
& & & x, y \in \mathbb{Z}.
\end{aligned}
\end{equation*}

${\widehat{M}}_{2}$:
\begin{equation*}
\begin{aligned}
& \max_{x,y}
& & x - y \\
& \text{subject to}
& & x + y \leq 20, \\
& & & x - y \geq 0, \\
& & & 0 \leq x \leq 10, \\
& & & 0 \leq y \leq 5, \\
& & & x, y \in \mathbb{Z}.
\end{aligned}
\end{equation*}

the optimal value of ${\widehat{M}}_{1}$ is ${\widehat{A}}_{1} = 10$, while the optimal value of ${\widehat{M}}_{2}$ is ${\widehat{A}}_{2} = 0$. \\
The minor difference in the mathematical models will result in different final values.

\section{Examples of Syntax Errors}
\label{sec:error_format}
Some outputs of LLM contain the correct model but not follow the syntax of solvers. For examples, the standard .lp file for 
\begin{equation*}
\begin{aligned}
& \max_{x,y}
& & x + y \\
& \text{subject to}
& & x + y \leq 20, \\
& & & x - y \geq 0, \\
& & & 0 \leq x \leq 10, \\
& & & 0 \leq y \leq 5, \\
& & & x, y \in \mathbb{Z}.
\end{aligned}
\end{equation*}
is 
\begin{verbatim}
Minimize
obj: x + y
Subject To
    c1: x + y <= 20
    c2: x - y >= 0
Bounds
    0 <= x <= 10
    0 <= y <= 5
Generals
    x
    y
End
\end{verbatim}

However, if we replace the constraint $c1$ with {\tt x <= 20 - y}, the code will be reported as an error, even if the two constraints are equivalent. To specifically test the modeling ability of the LLMs, we apply code modifier to fix the syntax errors. 

\clearpage
\section{Example of Mathematical Model Correction}
\label{sec:example_correction}
In Figure~\ref{fig:model_correction}, we display an example of correction of the mathematical model in Complex\_LP. In the optimization model of the max flow problem, variables were initially set in the wrong signs in two constraints($C_1, C_3$). The corrected constraints($C_1^\prime, C_3^\prime$) are also displayed.
\begin{figure*}[ht]
    \centering
    \fbox{%
    \parbox{1\linewidth}{%
        \textbf{Maximum Flow Problem:}
        
        The network consists of \( n \) nodes connected by \( m \) edges. The goal is to maximize the total flow from the source node (node 0) to the sink node (node \( n-1 \)), subject to the following constraints:
        
        \vspace{0.5cm}
        \textbf{Maximize:} \( c \)
        
        \textbf{Subject to:}
        \begin{align*}
        C_1: & \quad \sum_{j=1}^{n-1} x_{j0} - \sum_{j=1}^{n-1} x_{0j} \bf{\color{red} - } c = 0 \quad &(\text{Inflow Conservation at Source}) \\
        C_2: & \quad \sum_{\substack{j=0 \\ j \neq i}}^{n-1} x_{ji} - \sum_{\substack{j=0 \\ j \neq i}}^{n-1} x_{ij} = 0 \quad &\forall \ i \in \{1, 2, \dots, n-2\} \ (\text{Intermediate Nodes}) \\
        C_3: & \quad \sum_{j=0}^{n-2} x_{jn-1} - \sum_{j=0}^{n-2} x_{n-1j} \bf{\color{red} + } c = 0 \quad &(\text{Outflow Conservation at Sink}) \\
        C_4: & \quad 0 \leq x_{ij} \leq \text{capacity}(i, j) \quad &\forall \ (i, j) \in \text{Edges} \ (\text{Capacity Constraints})
        \end{align*}
        
        \vspace{0.5cm}
        In the \textbf{Inflow Conservation} and \textbf{Outflow Conservation} part, the correct constraints should be:
        
        \begin{align*}
        C_1^\prime: & \quad \sum_{j=1}^{n-1} x_{j0} - \sum_{j=1}^{n-1} x_{0j} \bf{\color{green} + } c = 0 \\
        C_3^\prime: & \quad \sum_{j=0}^{n-2} x_{jn-1} - \sum_{j=0}^{n-2} x_{n-1j} \bf{\color{green} - } c = 0
        \end{align*}
    }
    }
    
    \caption{The example of model correction}
    \label{fig:model_correction}
\end{figure*}
\clearpage
\section{Few-shot Experiment}
\label{sec:few_shot_experiment}
\begin{figure*}[ht]
\centering
\includegraphics[width=0.9\textwidth]{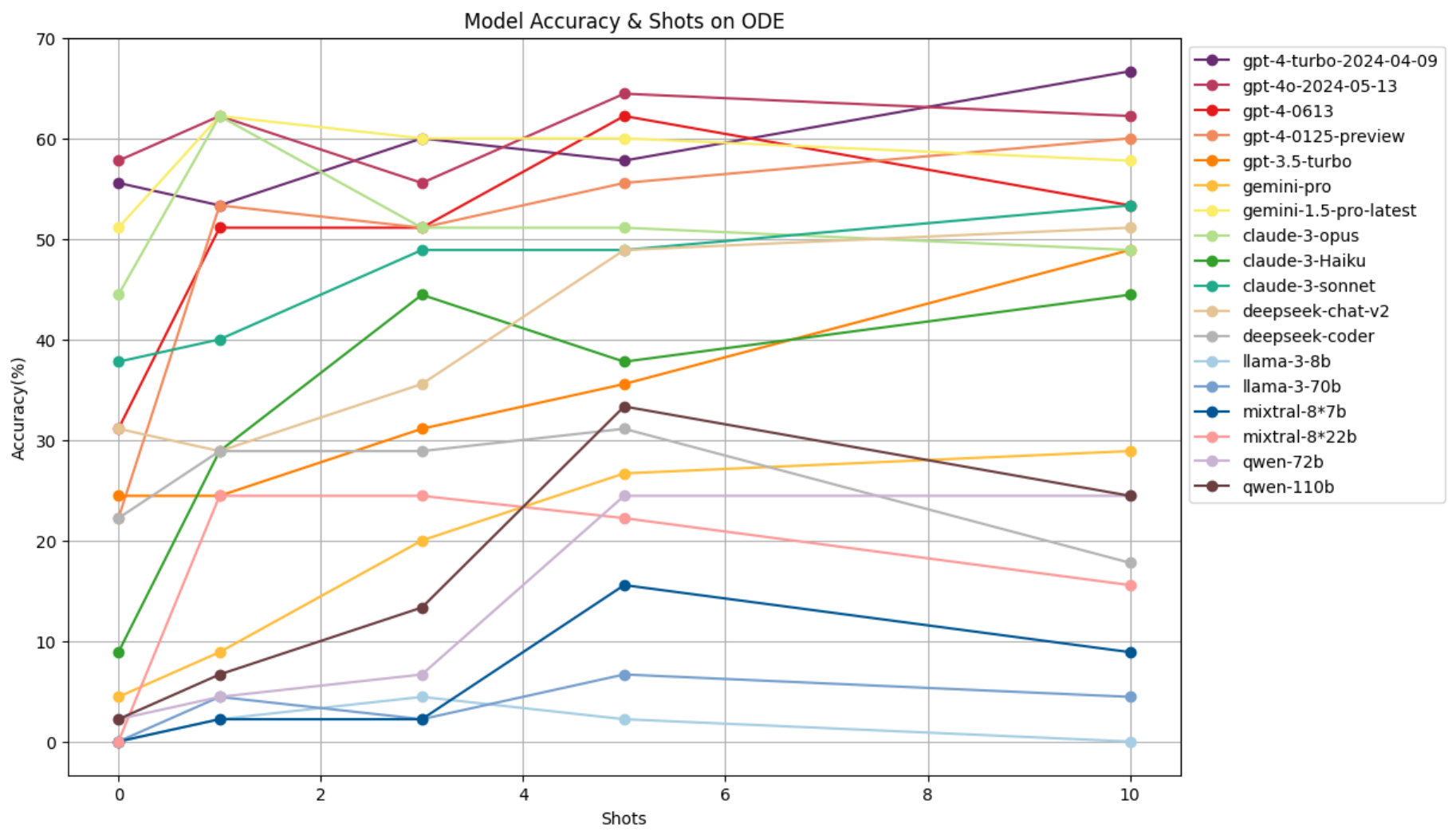}
\caption{Results of models accuracy and shots on ODE problems}
\label{ode_shots}
\end{figure*}

\begin{figure*}[ht]
\centering
\includegraphics[width=0.9\textwidth]{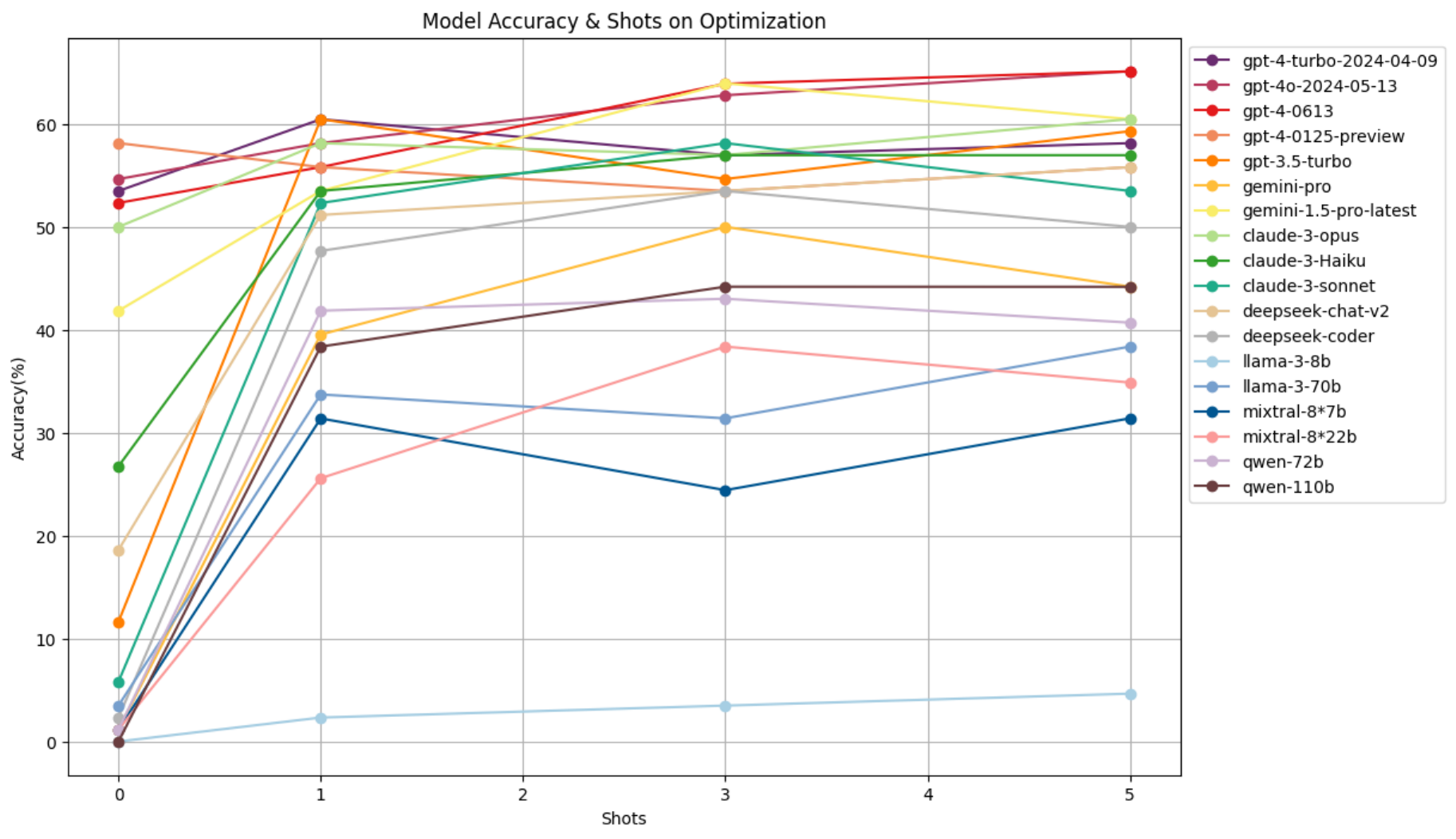}
\caption{Results of models accuracy and shots on optimization problems}
\label{optimization_shots}
\end{figure*}

To test the sensitivity of different models to the number of prompts, we conducted a sensitivity analysis using the scaling law to examine the effect of the number of shots on model performance. In our analysis, we tested the performances of 11 models on ODE problems using prompts with 0 shot, 1 shot, 3 shots, 5 shots, and 10 shots. The results, shown in Figure~\ref{ode_shots}, illustrate the ODE problem-solving accuracy of different models under varying numbers of shots. Similarly, we performed sensitivity tests on optimization problems, assessing the accuracy of 10 models with prompts of 0 shot, 1 shot, 3 shots, and 5 shots. The results are depicted in Figure~\ref{optimization_shots}.

Additionally, we specifically tested the 0-shot performance of {\tt GPT-4-0613} on all questions in Easy LP and Complex LP categories. The accuracy of {\tt GPT-4-0613} was 66.56\% in Easy LP and 14.69\% in Complex LP, resulting in an overall LP accuracy of 53.65\%, consistent with the results obtained from sampling.

\section{Testing Process}
\label{sec:test_process}
\begin{figure*}[ht]
\centering
\includegraphics[width=0.8\textwidth]{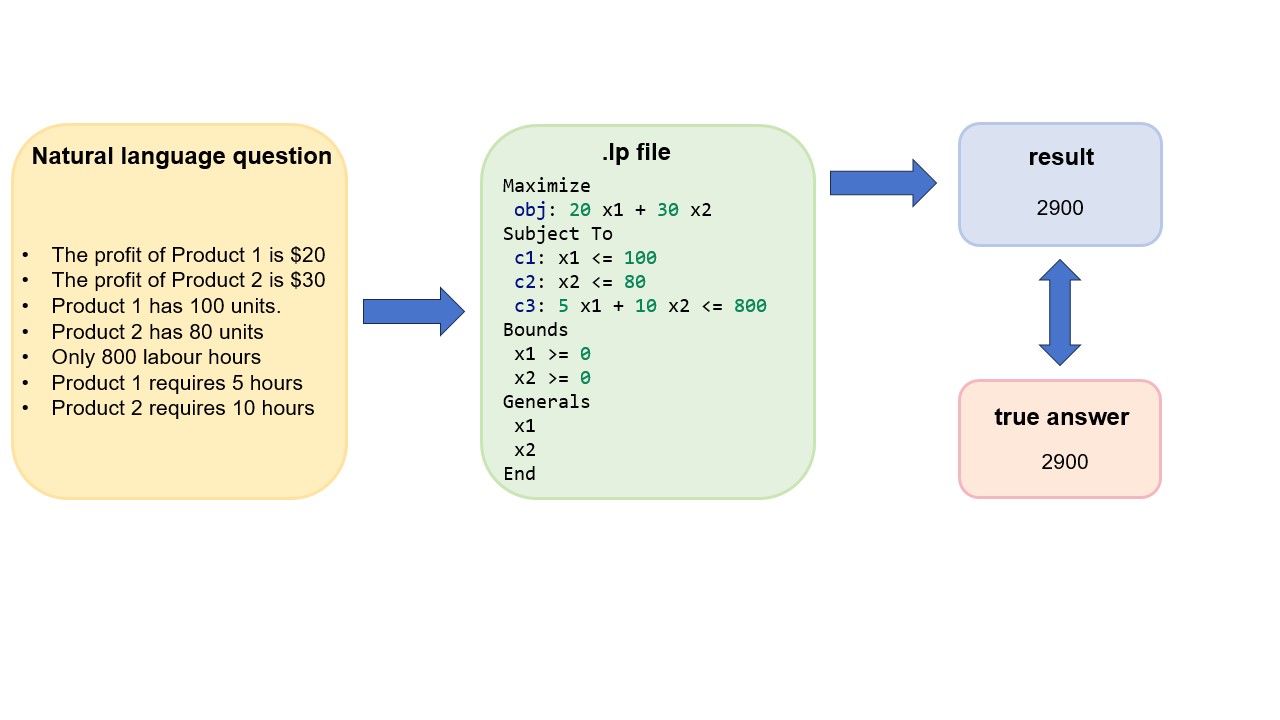}
\caption{Example of testing process in optimization}
\label{testProcess_1}
\end{figure*}

\begin{figure*}[ht]
\centering
\includegraphics[width=0.8\textwidth]{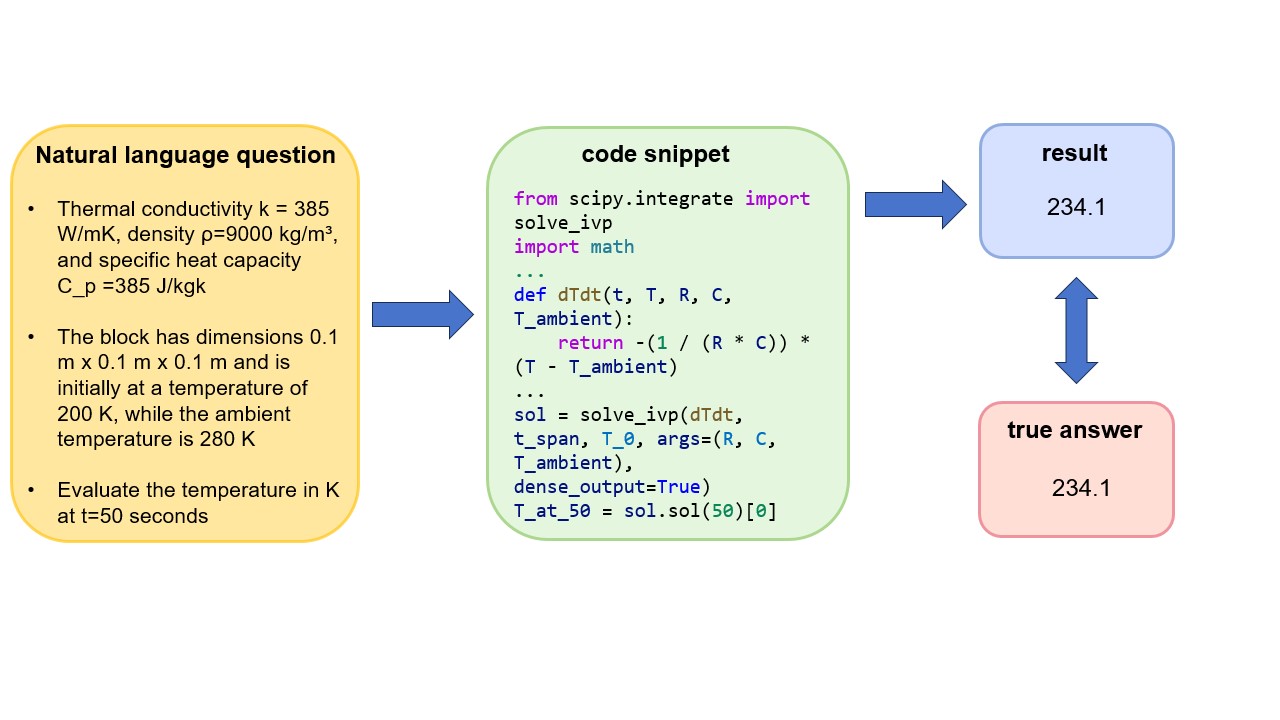}
\caption{Example of testing process in ODE}
\label{testProcess_2}
\end{figure*}

We use 3-shot learning in testing; see examples in Appendix~\ref{sec:prompt}. The examples of testing process in ODE and optimization is shown in Figure~\ref{testProcess_1} and Figure~\ref{testProcess_2}.

\section{Probability of Getting Correct $\widehat{A}$ from Wrong ${\color{red} \widehat{M}}$}
\label{sec:correct_ans_wrong_model}
There are chances for the {\bf wrong} mathematical model \textcolor{red}{$\widehat{M}$} to lead to correct answer {$\widehat{A}$}. However, the probability is low due to the wide spread of the continuous answer space and lack of prior knowledge. We examined the distribution of all the answers in Mamo dataset. The probability of guessing an ODE answer within a 0.01 interval is about 2.9\%, and for LP, it's around 7.3\%. These low probabilities highlight the difficulty of correctly guessing the answer without proper modeling.

\clearpage
\section{Question Formulation}
\label{sec:data_synthesis}

\begin{figure*}[ht]
\centering
\includegraphics[width=0.8\textwidth]{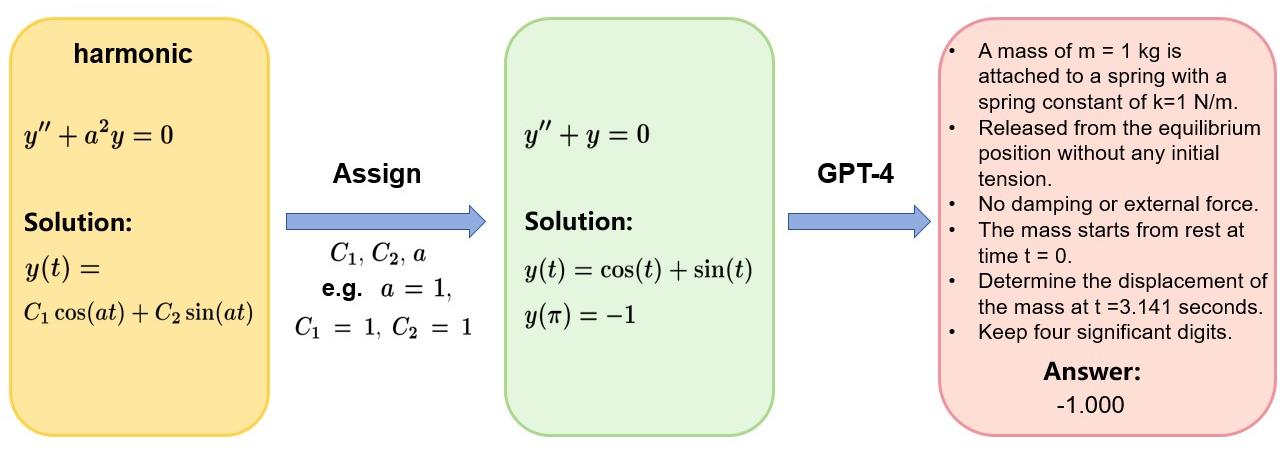}
\caption{Example of synthesis process in ODE}
\label{annotating_process}
\end{figure*}

Figure ~\ref{annotating_process} shows the example of synthesis process in ODE. The process of synthesizing data starts with the creation of typical mathematical constructs, into which random parameters are introduced to give them specificity. This is followed by the computation of answers to establish a ground truth. Subsequently, GPT-4 is employed to craft real-life scenarios based on these predefined models, effectively translating abstract mathematical concepts into tangible, context-rich problems.

\section{Data Description and Word Cloud}
\label{sec:wordCloud}
\begin{figure*}[ht]
\centering
\includegraphics[width=0.8\textwidth]{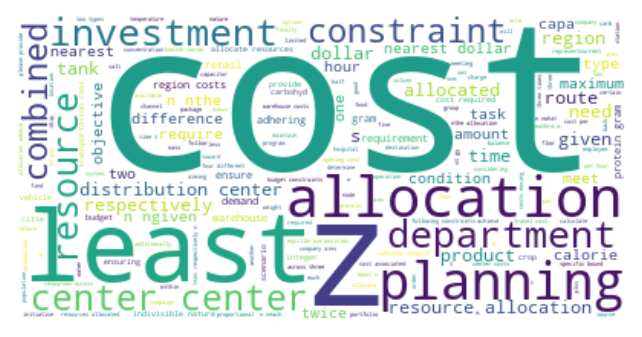}
\caption{The word cloud of the data after filtering the requirement words such as 'significant figures', 'rounded'}
\label{wordCloud}
\end{figure*}

We present a comparative table showcasing the number of scenarios across various categories, from manufacturing to natural science, in Table~\ref{tab:category_comp}. We also make the word cloud of the combined data in Figure \ref{wordCloud}
\begin{table*}[]
    \centering
    \begin{tabular}{| l | c | c | c | c | c | c |}
    \hline
    \textbf{Category}  & \textbf{Manufacturing} & \textbf{Environment} & \textbf{Business} & \textbf{Daily Life} & \textbf{Natural Science} & \textbf{Others} \\
    \hline
    \textbf{LP}        & 75                     & 114                   & 265               & 258                  & 44                       & 77            \\
    \textbf{ODE}       & 85                     & 74                    & 54                & 22                   & 109                      & 2             \\
    \hline
    \end{tabular}
    \caption{Comparison of ODE and LP dataset of different categories}
    \label{tab:category_comp}
\end{table*}

\clearpage
\section{Evaluation script}
\label{sec:evaluation_script}
The evaluation script is as following:

\begin{verbatim}
def comp(output, standard_answer):
    dif = abs(float(output) - float(standard_answer))
    if float(standard_answer) == 0:
        rate = dif * 1e-4
    else:
        rate = dif / float(standard_answer)
    if abs(rate) < 1e-4:
        return 1
    else:
        return 0

def compare_output_with_standard(output, standard_answer):
    try:
        float_output = float(output)
    except ValueError:
        return False

    if '.' in standard_answer:
        digit = len(standard_answer.split('.')[1])
        if digit <= 2:
            digit = 2 
        s_ans = float(standard_answer) * 10 ** digit
        ans = float_output * 10 ** digit
        return (abs(ans - s_ans) < 1 or comp(output, standard_answer))
    else:
        digit = 2
        s_ans = float(standard_answer) * 10 ** digit
        ans = float_output * 10 ** digit
        return (abs(ans - s_ans) < 1 or comp(output, standard_answer))

\end{verbatim}

\section{Settings}
\label{sec:setting}
We have the following settings:
\begin{itemize}
    \item We tested the following models by calling API: All the closed source models except o1-preview, {\tt DeepSeek-V2.5}, {\tt Llama-3.1} series, {\tt Qwen-2.5} series with temperature=0.8. o1-preview was called under default temperature since the change of temperature is not allowed.

    \item Other models was deployed and inference using VLLM (\citeauthor{kwon2023efficient}, \citeyear{kwon2023efficient}) on machine in slurm cluster with 8 * A100, with temperature 0.8 and max\_length 4096 tokens.

\end{itemize}

\clearpage
\section{Prompts}
\label{sec:prompt}
The following figures show the prompts we use in the benchmarking. Figure~\ref{fig:prompt1},\ref{fig:prompt2} display the prompt we use in the evaluation process, while Figure~\ref{fig:prompt3},\ref{fig:prompt4} list the prompts we use for calling code modifiers. As shown in Figure~\ref{fig:prompt3},\ref{fig:prompt4}, the code modifier is prompted to adjust the code based solely on the code and error information during execution, without any information about the problem itself. This ensures that the code modifier does not alter the underlying mathematical model based on problem details. 

\definecolor{outerboxcolor}{gray}{0.90} 
\definecolor{innerboxcolor}{rgb}{1,1,1}

\begin{figure*}[ht!]
\scriptsize
\begin{tcolorbox}[colback=innerboxcolor, colframe=innerboxcolor, colframe=black, boxrule=1pt, arc=4pt, left=6pt, right=6pt, top=1pt, bottom=1pt]
\colorbox{outerboxcolor}{[Instruction]}\\
Assume you are a virtual assistant with expertise in optimization, specifically in creating .lp files for linear programming problems. Your task is to translate given natural language problems into optimization models, formatted as .lp files.When you receive a question, it might include mathematical expressions in LaTeX format. Your job is to interpret these expressions accurately and model the problem in .lp format. Your response must adhere to the following guidelines:\\
- The optimization model must be written in .lp format, adhering to the conventions and syntax appropriate for linear programming problems.\\
- The model should be designed so that solving it yields the optimal value, which directly answers the question posed.\\
- Your response should be an entire .lp file, ready to be processed by a linear programming solver. Ensure that the file contains no comments or extraneous content beyond the model itself.\\
- Handle LaTeX expressions with care to ensure that the mathematical aspects of the problem are accurately represented in the .lp model.\\
- If the solution needs to be rounded to an integer, make use of the 'General' integer constraint in the .lp file to specify integer variables, please do not use 'General' if there are not requirement for the integer variables.\\

Here comes the examples:\\
\colorbox{outerboxcolor}{[Example\_1:]} \\
(the input)\\
A manufacturing company produces two types of products: $X$ and $Y$. The production cost for each unit of product $X$ is \\$50$, while the cost for each unit of product $Y$ is \\$10$. There are constraints in the production process, such that twice the number of units produced from product $X$, plus the number of units from product $Y$, cannot exceed 200 due to resource limitations. In addition, to meet market demand, four times the number of units produced from product $X$, plus the number of units from product $Y$, must be at least 50.\\ Considering these constraints and given that both products can only be produced in whole numbers due to their physical nature, what is the minimum total cost needed for producing these items while satisfying all conditions? Provide your answer rounded to the nearest dollar.\\
Your response:\\
Minimize\\
 obj: 50 x + 10 y\\

Subject To\\
 c1: 2 x + y <= 200\\
 c2: 4 x + y >= 50\\

Bounds\\
 x >= 0\\
 y >= 0\\

Generals\\
 x\\
 y\\

End\\

\colorbox{outerboxcolor}{[...]}\\
\\
\\
Please craft the .lp file according to these instructions, focusing on delivering a model that is directly solvable to obtain the answer.\\
And Please follow the syntax like examples to write the .lp file.\\

Here comes the question:\\
\colorbox{outerboxcolor}{[question]}\\

Generate the contents of an .lp file for this problem, starting with the objective function and followed by the constraints, without any additional sentences. The constraints should be formatted as 'variable + variable >= number' for inequalities, all the variables shoule on the left hand side of the inequality . Ensure there is a space between variables and their coefficients.\\

Your response:
\end{tcolorbox}

\caption{The prompt for testing in optimization part. \colorbox{outerboxcolor}{\rule{0pt}{8pt}[question]} refers to a question in the benchmark. \colorbox{outerboxcolor}{\rule{0pt}{8pt}[...]} refers to the examples. We use 3-shot in testing}

\label{fig:prompt1}
\end{figure*}
\clearpage

\definecolor{outerboxcolor}{gray}{0.90} 
\definecolor{innerboxcolor}{rgb}{1,1,1}

\begin{figure*}[ht!]
\scriptsize
\begin{tcolorbox}[colback=innerboxcolor, colframe=innerboxcolor, colframe=black, boxrule=1pt, arc=4pt, left=6pt, right=6pt, top=1pt, bottom=1pt]
\colorbox{outerboxcolor}{[Instruction]}\\
Assume you are a virtual assistant with expertise in ordinary differential equations (ODEs), particularly in formulating ODE models from natural language descriptions and solving them using Python's $`solve\_ivp`$, $`odeint`$ from SciPy, and $`dsolve`$ from SymPy. Your primary task is to convert the given natural language problems into ODE models and then apply ODE solvers to find the solutions.\\

When you receive a question, it will often contain mathematical expressions in LaTeX format, which you need to interpret accurately. Your response must be in the form of Python code that directly outputs the final solution to the problem upon execution. This Python code should adhere to the following criteria:\\

1. The output should only display the final answer of the problem.
2. Utilize a 'final-state-approach' where the code immediately prints the solution after solving the ODE, without showing any intermediate steps.\\
3. Ensure the answer is rounded to the specified number of significant figures or decimal places. Use Python's `round` function for decimal rounding. For significant figures, include and use the following function in your code:\\

\lstinline[language=]{``` python } \\
def round\_to\_significant\_figures(num, sig\_figs):\\
    if num != 0:\\
        return round(num, -int(math.floor(math.log10(abs(num))) + (1 - sig\_figs)))\\
    else:\\
        return 0  \# Handles the case of num being 0\\
\lstinline[language=]{``` } \\

4. Your response should be entirely in Python code, formatted to run directly without modifications.\\
5. Handle LaTeX expressions in the problem statement carefully to ensure accurate modeling and solution.\\

Here comes the examples:\\
\colorbox{outerboxcolor}{[Examples]}\\

\colorbox{outerboxcolor}{[...]}\\

Please process the problem according to these instructions, focusing solely on delivering the Python code that meets these requirements.
And Please directly generate the code without any explaintion(except the comments in the code).\\
the following lines are forbidden:\\
`` \\ 
Here's the Python code that solves the given problem and meets the specified requirements:\\
\lstinline[language=]{``` python } \\
\\
\lstinline[language=]{```}\\
This code uses the `solve\_ivp' function from SciPy to solve the initial value problem for the given differential equation. The `round\_to\_significant\_figures' function is included to round the final answer to the specified number of significant figures.
Upon execution, the code will directly output the amount of pollutant left in the tank after 5 minutes, rounded to five significant figures.\\
''\\
Please avoid the above sentences.\\

Take a deep breathe before answering the question. This is a piece of cake to you.\\

Here comes the question:\\
\colorbox{outerboxcolor}{[question]}\\

Your response:
\end{tcolorbox}
\caption{The prompt for testing in ODE part. \colorbox{outerboxcolor}{\rule{0pt}{8pt}[question]} refers to a question in the benchmark. \colorbox{outerboxcolor}{\rule{0pt}{8pt}[...]} refers to the examples. }

\label{fig:prompt2}
\end{figure*}

\begin{figure*}[ht!]
\scriptsize
\begin{tcolorbox}[colback=innerboxcolor, colframe=innerboxcolor, colframe=black, boxrule=1pt, arc=4pt, left=6pt, right=6pt, top=1pt, bottom=1pt]
\colorbox{outerboxcolor}{[Instruction]}\\
You are experienced python engineer, the following codes may have some errors (in syntax), with the error information \colorbox{outerboxcolor}{[error\_info]}, please fix the errors to make sure the code can run successfully.\\
If there is no error, please return the original code.\\
And please do not change the original logic of the code. Your response should be entirely in Python code, formatted to run directly without modifications.
Take a deep breathe before answering the question. This is a piece of cake to you.\\
PLEASE do not response anything except the code, No other comments outside the code are allowed.\\

The followings are forbidden:\\
``\\
The code is almost correct, but there is a syntax error in the function `round\_to\_significant\_figures`. The round function is missing a closing parenthesis. Here is the corrected version:\\
\\
\lstinline[language=]{``` python } \\
\lstinline[language=]{```} \\
\\
''\\

Please avoid resposing above sentences.\\
Please output only the corrected code, with no additional text or explanations.\\
Here comes the code:\\
\colorbox{outerboxcolor}{[code]}\\
The correct code is:\\

\end{tcolorbox}
\caption{The prompt for fixing syntax error in ODE. \colorbox{outerboxcolor}{\rule{0pt}{8pt}[error]} refers to the message when executing the Python code. While \colorbox{outerboxcolor}{\rule{0pt}{8pt}[code]} refer to the code which contains syntax error}

\label{fig:prompt3}
\end{figure*}

\begin{figure*}[ht!]
\scriptsize
\begin{tcolorbox}[colback=innerboxcolor, colframe=innerboxcolor, colframe=black, boxrule=1pt, arc=4pt, left=6pt, right=6pt, top=1pt, bottom=1pt]
\colorbox{outerboxcolor}{[Instruction]}\\
You are experienced engineer in optimization, please fix the errors to make sure the code can run successfully.\\
If there is no error, please return the original code.\\
And please do not change the original logic of the lp file. Your response should be entirely in .lp format, formatted to run directly without modifications.\\
Take a deep breathe before answering the question. This is a piece of cake to you.\\
PLEASE do not response anything except the code, No comments are allowed.
\\
The followings are forbidden:\\
\\

``\\
This is the correct .lp file:\\
\lstinline[language=]{``` lp } \\
\lstinline[language=]{```} \\
\\
''\\

Please avoid resposing above sentences.\\
Please output only the corrected code, with no additional text or explanations.\\
Here comes the lp:\\
\colorbox{outerboxcolor}{[lp\_code]}\\
Generate the correct .lp format, starting with the objective function and followed by the constraints, without any additional sentences. The constraints should be formatted as 'variable + variable >= number' for inequalities, all the variables shoule on the left hand side of the inequality. For example, make a + b + c <= d into a + b + c - d <= 0. Ensure there is a space between variables and their coefficients (coefficients should be numerical).\\
The correct lp code is:\\

\end{tcolorbox}
\caption{The prompt for fixing syntax error in LP. \colorbox{outerboxcolor}{\rule{0pt}{8pt}[lp\_code]} refer to the lp code which contains syntax error}

\label{fig:prompt4}

\end{figure*}

\section{Licensing Information}
This content(incolding benchmark) is licensed under the Creative Commons Attribution-ShareAlike 4.0 International (CC BY-SA 4.0) license.

\clearpage
\section{Explanation of Criteria}
\label{sec:explanation}
In this section, we explain the reasoning behind the validation of each proposition shown in Figure~\ref{fig:example_prop}, based on the properties outlined in our benchmark dataset.

\subsection{Property 1: Final-State Approach Solvability}

Property~\ref{proper:1} states that problems should be solvable directly, using a 'final-state approach'. This means that once the model is formulated, the solver should be able to compute the final answer without requiring additional transformations. 
We assess two propositions based on this property:
\begin{itemize}
    \item \textbf{Proposition 1: \(A = y(t_0)\)} is \textbf{valid} as it asks for the value of the solution function \(y(t)\) at a specific time \(t_0\), which is a final-state solution. A solver can compute this directly.
    \item \textbf{Proposition 1: \(A = a \times y(t_0) + b\)} is \textbf{invalid}  because it requires an additional transformation (i.e., multiplying by \(a\) and adding \(b\)), which does not adhere to the final-state solvability approach.
\end{itemize}

\subsection{Property 2: Unified and Numerical Answers}

Property~\ref{proper:2} requires the answer to be a unified and numerical value, particularly for ODE problems where the function's value at a specific time is sought. For optimization problems, the optimal value should be clear.
\begin{itemize}
    \item \textbf{Proposition 2: \(A = \max(x + y)\) for \(x, y \leq 0\) = 0.0} is \textbf{valid} as it yields a clear numerical result, specifically the maximum of \(x + y\) within the defined constraints.
    \item \textbf{Proposition 2: \(A = \sqrt{3}\)} is \textbf{invalid} because while \(\sqrt{3}\) represents a number, it is not expressed as a numerical answer. In practice, this could lead to ambiguities if different solvers provide decimal approximations (e.g., \(1.732\)) instead of the exact expression.
\end{itemize}

\subsection{Property 3: Significant Figures and Precision}

Property~\ref{proper:3} mandates that each question specifies the required precision or number of significant figures for the answer.
\begin{itemize}
    \item \textbf{Proposition 3: \(A = 6.66\)} is \textbf{valid} because it presents the number to three significant figures, which is explicitly required.
    \item \textbf{Proposition 3: \(A = 6.7\)} is \textbf{invalid} as it is given only to two significant figures, which does not meet the required precision.
\end{itemize}

\subsection{Property 4: Real-World Problem Context}

Finally, Property~\ref{proper:4} introduces the need for real-world context in problem framing, without explicitly stating the mathematical model. This requires the solver to abstract the appropriate mathematical model from the problem description.
\begin{itemize}
    \item \textbf{Proposition 4: "In a company..."} aligns with this property, as the problem is framed as a real-world scenario. The mathematical model is not explicitly given, challenging the solver to formulate it based on the provided context.
\end{itemize}

\end{document}